\documentclass[10pt,journal,compsoc]{IEEEtran}

\usepackage{times}
\usepackage{epsfig}
\usepackage{graphicx}
\usepackage{amsmath}
\usepackage{amssymb}
\usepackage{xcolor}
\usepackage{algorithm}
\usepackage{algpseudocode}
\usepackage{wrapfig}
\usepackage[nocompress,nospace, noadjust, sort]{cite}
\usepackage{enumitem}

\usepackage[pagebackref=false,breaklinks=true,letterpaper=true,bookmarks=false,backref=true]{hyperref}

\DeclareMathOperator*{\argmin}{arg\,min}
\DeclareMathOperator*{\argmax}{arg\,max}
\renewcommand\footnotemark{}

\begin{document}

\title{Context-Aware Query Selection for Active Learning in Event Recognition}

\author{Mahmudul Hasan$^*$$^1$, Sujoy Paul$^*$$^2$, Anastasios I. Mourikis$^2$, and Amit K. Roy-Chowdhury$^2$\\
	$^1$Comcast Labs, $^2$University of California, Riverside \\
	{\tt\small \{mhasa004@, spaul003@, mourikis@ee., amitrc@ee.\}ucr.edu}
\IEEEcompsocitemizethanks{\IEEEcompsocthanksitem Mahmudul Hasan was with the Dept. of Computer Science and Engineering at the University of California Riverside. Currently, he is a Senior Researcher at Comcast Labs, Washington, DC. Sujoy Paul, Anastasios I. Mourikis, and Amit K. Roy-Chowdhury are with the Dept. of Electrical and Computer Engineering at the University of California Riverside.
\IEEEcompsocthanksitem This work was supported in part by NSF under grant IIS-1316934, by the US Department of Defense, and by Google.
\IEEEcompsocthanksitem First two authors should be considered as joint first authors.}
\thanks{Manuscript received ****; revised August ****.}}

\markboth{Journal of Pattern Analysis and Machine Intelligence,~Vol.~*, No.~*, Month~Year}{}%

\IEEEtitleabstractindextext{
\begin{abstract}
Activity recognition is a challenging problem with many practical applications. In addition to the visual features, recent approaches have benefited from the use of context, e.g., inter-relationships among the activities and objects. However, these approaches require data to be labeled, entirely available beforehand, and not designed to be updated continuously, which make them unsuitable for surveillance applications. In contrast, we propose a continuous-learning framework for context-aware activity recognition from unlabeled video, which has two distinct advantages over existing methods. First, it employs a novel active-learning technique that not only exploits the informativeness of the individual activities but also utilizes their contextual information during query selection; this leads to significant reduction in expensive manual annotation effort. Second, the learned models can be adapted online as more data is available. We formulate a conditional random field model that encodes the context and devise an information-theoretic approach that utilizes entropy and mutual information of the nodes to compute the set of most informative queries, which are labeled by a human. These labels are combined with graphical inference techniques for incremental updates. We provide a theoretical formulation of the active learning framework with an analytic solution. Experiments on six challenging datasets demonstrate that our framework achieves superior performance with significantly less manual labeling.
\end{abstract}

\begin{IEEEkeywords}
	Active Learning, Activity Recognition, Visual Context, Information Theory.
\end{IEEEkeywords}}

\maketitle

\IEEEdisplaynontitleabstractindextext

\IEEEpeerreviewmaketitle

\section{Introduction}
\label{sec:intro}
\IEEEPARstart{H}uge amounts of video data are being generated nowadays from various sources, and can be used to learn activity recognition models for video understanding. Learning usually involves supervision, i.e., manually labeling instances and using them to estimate model parameters. However, there may be drift in concepts of activities and new types of activities can arise. In order to incorporate this dynamic nature, activity recognition models should be learned continuously over time, and be adaptive to such changes. However, manually labeling a huge corpus of data continuously over time is a tedious job for humans, and prone to anomalous labeling. To reduce the manual labeling effort, without compromising the performance of the recognition model, active learning~\cite{S12}  can be used.
\begin{figure}[t]

\centering
\includegraphics[scale=0.37]{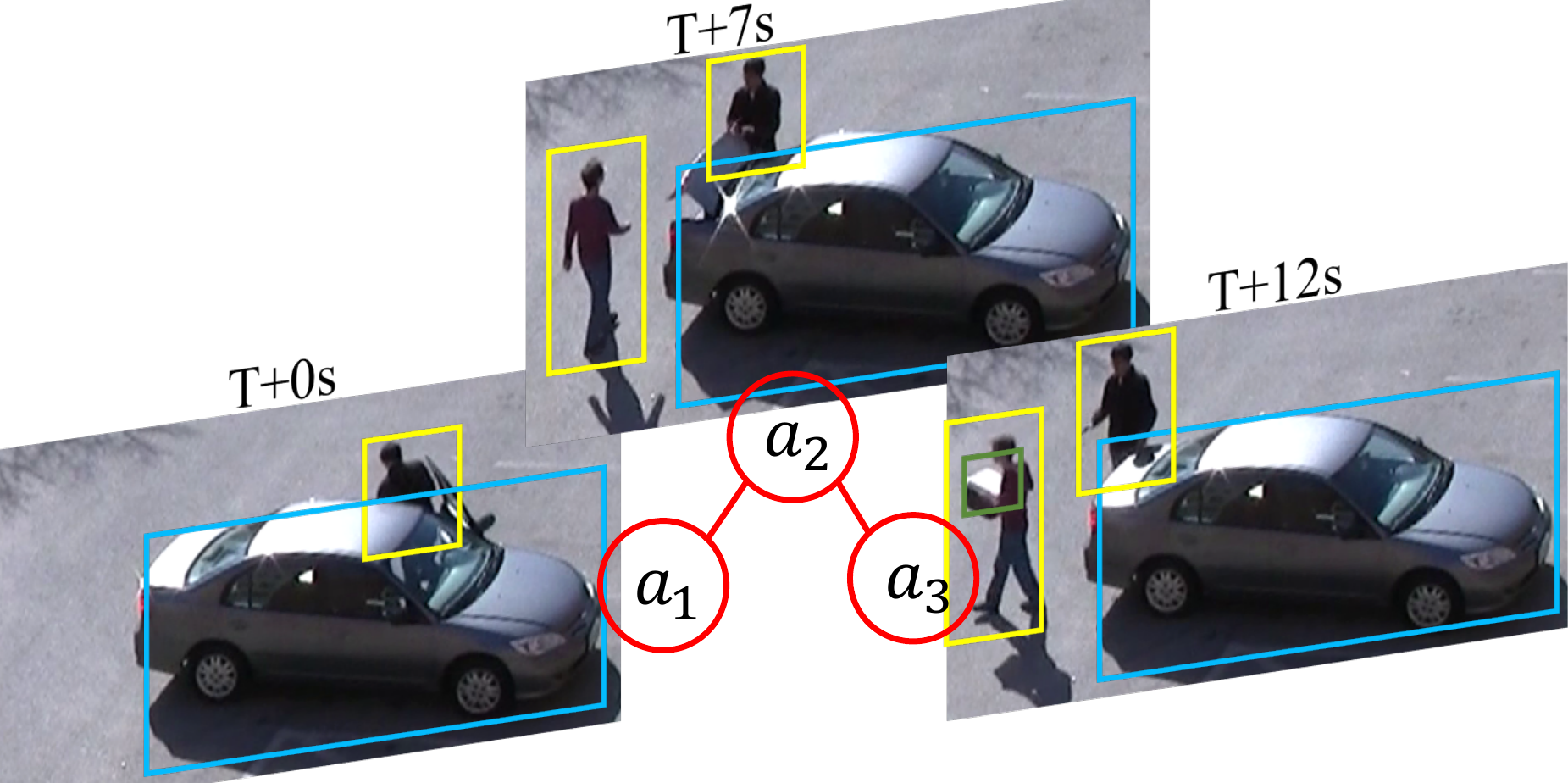}
\caption{A sequence of a video stream \cite{SHP+11} shows three new unlabeled activities - \textit{person getting out of a car} ($a_1$) at $T+0s$, \textit{person opening a car trunk} ($a_2$) at $T+7s$, and \textit{person carrying an object} ($a_3$) at $T+12s$. These activities are spatio-temporally correlated, and this information can provide context. Conventional approaches to active learning for activity recognition do not exploit these relationships in order to select the most informative instances. However, our approach exploits context and actively selects instances (in this case $a_2$) that provide maximum information about other neighbors.}
\label{fig:motiv}
\end{figure}

\begin{figure*}[ht]
\centering
\includegraphics[scale=0.65]{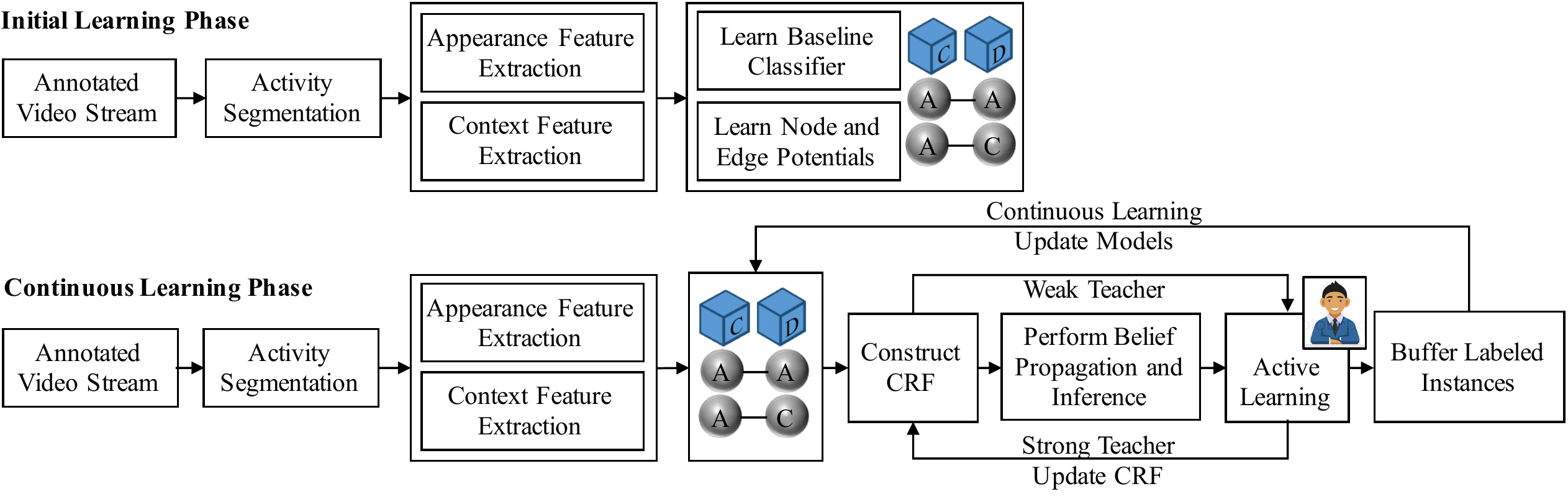}
\caption{Our proposed framework for learning activity models continuously. Please see the text in Section \ref{sec:overview} for details.}
\label{fig:framework}
\end{figure*}

Several visual-recognition systems exploit the co-occurrence relationships between objects, scene, and activities that exist in natural settings. This information is often referred to as {\em context}~\cite{OT07}.
Human activities, in particular, are not only related spatially and temporally, but also have relationships with the surroundings (e.g. objects in the scene), and this information can be used for improving the performance of recognition models. (Figure \ref{fig:motiv}). Several prior research efforts \cite{YF10,WSS13,LYWM10,ZNR13} have considered the use of context for recognizing human activities and showed significant performance improvement over context-free approaches. However, these context-aware approaches assume that large numbers of instances are manually labeled and available for training the recognition models. Although some methods \cite{RJS09,MMW12,HC14} learn human activity models incrementally from streaming videos, they do not utilize contextual information to select \textit{only} the informative samples for manual labeling, so as  to reduce human effort. In this work, {\em we develop an active-learning framework, which exploits the contextual relationships among activities and objects to reduce the manual labeling effort, without compromising recognition performance.}


Active learning has become an important tool for selecting the most informative instances from a large volume of unlabeled data to be labeled by a human annotator. In order to select the informative instances, most active learning approaches \cite{S12} exploit metrics such as informativeness, expected error reduction (EER), or expected change in gradients. Such criteria for query selection are based on {\em individual} data instances, and do not assume that relationships exist among them. However, as mentioned previously, activities and objects in video exhibit significant interrelationships, which can be encoded using graphical models, and exploited when selecting the most informative queries for manual labeling (see Figure \ref{fig:motiv}).

Related problems have been studied in other computer science areas.
For example, the authors in~\cite{shi2012batch} exploit link-based dependencies in a network-based representation of the data, while \cite{mac2014hierarchical} utilizes the interrelationship of the data instances in feature space for active learning. Some works \cite{settles2008analysis} perform query selection on a conditional random field (CRF) model for structured prediction in natural language processing by utilizing only the co-occurrence relationships that exist among the tokens in a sentence, while activities in a video sequence additionally exhibit spatial and temporal relationships as well as interactions with objects. Hence, it is a challenging task to select the informative samples by exploiting the interrelationships within the instances to reduce the manual labeling effort.

\subsection{Main Contributions and Overview}
\label{sec:overview}
In this work, we propose a novel active-learning framework that exploits contextual information encoded using a CRF, in order to learn an activity recognition model from videos. The {\bf main contribution} of this work is twofold:
\begin{enumerate}
	\item A new query-selection strategy on a CRF graphical model for inter-related data instances, utilizing entropy and mutual information of the nodes.
	\item Continuous learning of both the activity recognition and the context models simultaneously, as new video observations come in, so that the models can be adaptive to the changes in a dynamic environment.
\end{enumerate}
The above two contributions rely on a CRF model that is automatically constructed online, and can utilize any number and type of context features. An overview of our proposed framework is illustrated in Figure \ref{fig:framework}.

Our framework has two phases: initial learning phase and incremental learning phase. During the initial learning phase, with a small amount of annotated videos in hand, we learn a baseline activity classifier and spatio-temporal contextual relationships. During the incremental learning phase, given a set of unlabeled activities, we construct a CRF with two types of nodes: activity nodes and context nodes. Probabilities from the baseline classifier are used as the activity node potentials, while object detectors are used to detect context features and to compute the context node potentials. In addition to the contextual information encoded in the context nodes (termed {scene-activity} context), we also use inter-activity contextual information. This represents the co-occurrence relationships between activities, and is encoded by the edge potentials among the activity nodes. For recognition, we perform inference on the CRF in order to obtain the marginal probabilities of the activity nodes.

We propose a novel active learning framework, which leverages upon both a strong teacher (human) and a weak teacher (recognition system output) for labeling. We choose for manual labeling the activity nodes that minimize the joint entropy of the CRF. This entropy can be approximately computed using the entropy of the nodes and the mutual information between pairs of connected nodes. After acquiring the labels from the strong teacher (which is assumed to be perfect), we run an inference on the graph conditioned on these labeled instances. The labeled nodes help the unlabeled ones to improve the confidence in their classification decisions, i.e., reduce the entropy of their classification probability mass functions. The unlabeled nodes that attain high confidence after the inference are also included in the training set, and constitute the input of the weak teacher. The newly labeled instances are then used to update the classifier as well as the context models.

The work presented in this paper is a more comprehensive version of a previously published paper~\cite{hasan2015context}. In addition to a more detailed presentation and new experiments, new fundamental technical contributions are included in this paper, providing an improved framework for context-aware active learning. Specifically, in our previous work, we intuitively derived the query-selection strategy and only provided a greedy solution. In this   work, we derive an information-theoretic query-selection criterion from first principles, and provide a branch-and-bound solution with provable convergence properties. We conduct new experiments to show the effectiveness of our method and demonstrate that it outperforms the results in our previous work.

\section{Relation to Existing Works}
\label{sec:related}
Our work involves the following areas of interest: human activity recognition, active learning, and continuous learning. We here review relevant papers from these areas.

{\bf Activity recognition.}
Visual activity-recognition approaches can be classified into three broad categories: those using interest-point based low-level local features; those using human-track and pose-based mid-level features; and those using semantic-attribute-based high-level features. Survey article \cite{P10} contains a more detailed review on feature-based activity recognition. Recently, context has been successfully used for activity recognition. The definition of context may vary based on the problem of interest. For example, \cite{YF10} used object and human pose as the context for activity recognition from single images. Collective or group activities were recognized in \cite{LYWM10} using the context in the group. Spatio-temporal contexts among the activities and the surrounding objects were used in \cite{ZNR13}. Graphical models were used to predict human activities in \cite{WSS13}. However, most of these approaches are batch-learning algorithms that require all of the training instances to be present and labeled beforehand. On the contrary, we aim to learn activity models continuously from unlabeled data, with minimum human labeling effort.

{\bf Active learning.}
It has been successfully applied to many computer vision problems including tracking \cite{vondrick2011video}, object detection \cite{vijayanarasimhan2014large}, image \cite{batra2010icoseg} and video segmentation \cite{fathi2011combining}, and activity recognition \cite{LZ11}. It has also been used on CRFs for structured prediction in natural-language processing \cite{druck2009active, settles2008analysis}. These methods use information-theoretic criteria,  such as the entropy of the individual nodes, for query selection. We here follow a similar approach, but additionally model the mutual information between nodes, because different activities in video are related to each other. Our criterion captures the entropy in each activity, but subtracts the conditional entropy of that activity when some other related activities are known. As a result, our framework can select the most informative queries from a set of unlabeled data represented by a CRF.

Some prior research \cite{Chakraborty2015Active, elhamifar2013convex} has considered active learning as a batch-selection problem, and has proposed convex relaxations of the resulting non-convex formulations. The work in \cite{sun2015active} performs active learning on a CRF and provides a solution to the exact, computationally intractable, problem by histogram approximation of   Gibbs sampling. Methods in \cite{Chakraborty2015Active} and \cite{elhamifar2013convex} perform active learning for the image labeling problem, where interrelationships are measured by the KL-divergence of the class probability distribution of similar neighboring instances. The method in \cite{sun2015active} performs active learning for image segmentation by only considering the spatial relationships among the neighboring super pixels. On the contrary, our active-earning system can take the advantage of both spatial and temporal relationships among the activities and context attributes in the video sequence.

{\bf Continuous learning.}
Among several schemes on continuous learning from streaming data, methods based on an ensemble of classifiers \cite{HSK+11} are most common. In these, new weak classifiers are trained with the newly available data and added to the ensemble. Only few methods can be found that learn activity models incrementally. The feature-tree-based method proposed in \cite{RJS09} grows in size with new training data. The method proposed in \cite{MMW12} uses human tracks and snippets for incremental learning. The most closely related works are \cite{HC14} and \cite{hasan2016incremental}, which are based on active learning and boosted SVM classifiers. However, the approach of \cite{HC14} does not exploit contextual relationships, while \cite{hasan2016incremental} does not take advantage of the mutual information among the activity instances. In this work, we exploit both context attributes and mutual information, thereby increasing recognition performance, while keeping the human labeling cost small.

\section{Modeling Contextual Relationships}
\label{sec:modeling}

Let us consider that after segmenting a video stream, we obtain $n$ activity instances (segments) to be recognized. These activity instances may occur at different times in the stream, or at different spatial locations simultaneously. We denote these activity instances as $a_i $, $i=1,\hdots, n$, and the set of all possible activity classes, in which these activities belong, is denoted as $\mathcal S_a = \{A_1, \hdots, A_q$\}.
We assume that a ``baseline'' activity-recognition model $\mathcal P$ is available, that can be used to obtain the prior class-membership probabilities for each of the activities $a_i$. This baseline model operates using features extracted in the video; we denote the features extracted in the $i$-th activity segment for use by $\mathcal P$ as $x_i \in \mathcal S_x$, where $\mathcal S_x$ is the space of activity features.

As mentioned earlier, we employ two types of contextual attributes, namely scene-activity and inter-activity attributes. Intra-activity context attributes are scene-level features and object attributes related to the activity of interest, whereas inter-activity context represents relationships among the neighboring activities. These context attributes are not low-level features, but may provide important and distinctive visual clues. We denote both of these context attributes as $\mathcal C$.  

We assume that a set of detection algorithms $\mathcal D$ is available, which operate on low-level image data in the $i$-th activity segment, $z_i \in \mathcal S_z$, to compute the prior probability for the contextual attributes of this segment, $c_i \in \mathcal S_c$. Depending on the specific application, we may use several different types of attributes. If, for example, two types of attributes are used, then $c_i = [c_i^1, c_i^2]$, with $c_i^1 \in  \mathcal S_c^1$ and $c_i^2 \in \mathcal S_c^2$, and $\mathcal S_c  = \mathcal S_c^1 \times \mathcal S_c^2$. Specific examples of contextual attribute types are provided later on in this section.

We formulate a generalized CRF model for activity recognition, that does not depend on any particular choice of feature extraction algorithms, baseline classifiers, or context attributes. In Section \ref{sec:expt}, we describe the specific choices we made during our experiments.

\begin{figure}[t]
\centering
\includegraphics[scale=0.7]{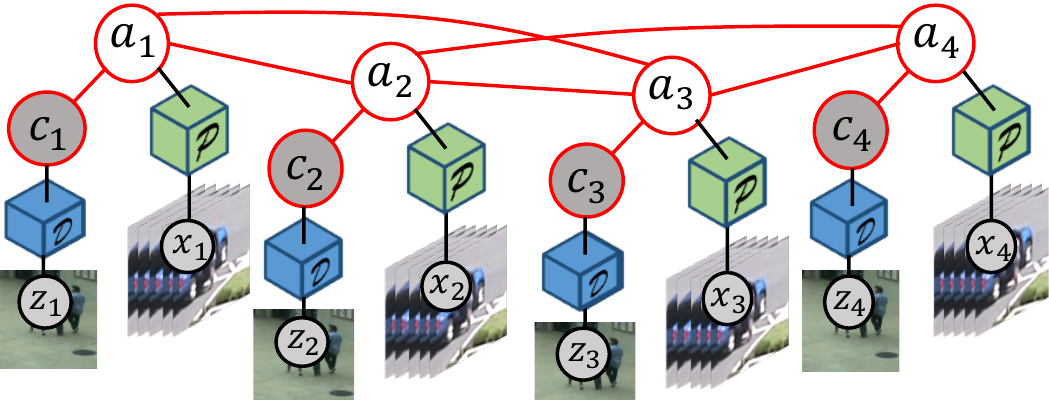}
\caption{Illustration of a CRF for encoding the contextual information. Please see the text in Section \ref{sec:modeling} for details.}
\label{fig:crf}
\end{figure}

{\bf Overview.} We model the interrelationships among the activities and the  context attributes  using a CRF graphical model as shown in Figure \ref{fig:crf}. This consists of an undirected graph $G=(V,E)$ with a set of nodes $V = \{V_a,V_c,V_x,V_z\},$ and a set of edges $E = \{V_a-V_a,V_a-V_c,V_a-V_x,V_c-V_z\}$. Here $V_a = \{a_i\}_{i=1}^{n}$ are the activity nodes, $V_c = \{c_i\}_{i=1}^{n}$ are the context attribute nodes, and $V_x = \{x_i\}_{i=1}^{n}$ and $V_z = \{z_i\}_{i=1}^{n}$ are the observed visual features for the activities and the context respectively.  In Figure \ref{fig:crf}, $\mathcal{P}$ represents the activity classifier and $\mathcal{D}$ stands for the object detectors. They are used to compute the prior node potentials and to construct the context features respectively. We are interested in computing the posterior of the $V_a$ nodes. Red edges among the $V_a$ and $V_c$ nodes represent spatio-temporal relationship among them. The connections between $V_a$ and $V_c$ nodes are fixed but we automatically determine the connectivity among the $A$ nodes along with their potentials. The overall potential function ($\Phi$) of the CRF is shown in Equation \ref{eqn:energy}, where $\phi$s and $\psi$s are node and edge potentials. We define the potential functions as follows.
\begin{align}
&\Phi  = \prod_{\substack{a_i\in V_a, c_i \in V_c\\x_i \in V_x, z_i\in V_z}}\phi(a_i,x_i) \phi(c_i,z_i) \prod_{\substack{a_i,a_j \in V_a\\c_i\in V_c}} \psi(a_i,a_j) \psi(a_i,c_i) \label{eqn:energy}\\
&\phi(a_i,x_i) = p(a_i|x_i, \mathcal{P}) \label{eqn:nodeAPot}\\
&\phi(c_i,z_i) = \phi(c_i^1,z_i) \odot \phi(c_i^2,z_i) \label{eqn:contextPot}\\
&\phi(c^1_i,z_i) = p(c^1_i|z_i, \mathcal{D})\label{eqn:objectAttribute}\\
&\phi(c^2_i,z_i) = \text{bin}(c_i^2)~\mathcal{N}(c_i^2, \mu_{c^2}, \sigma_{c^2})\label{eqn:peopleAttribute}\\
&\psi(a_i,a_j) = F_a(a_i,a_j)~ \mathcal{N}(\|t_{a_i}-t_{a_j}\|^2,\mu_t,\sigma_t) \notag\\ &\qquad\qquad~~~~\mathcal{N}(\|s_{a_i}-s_{a_j}\|^2,\mu_s,\sigma_s) \label{eqn:edgeAAPot}\\
&\psi(a_i,c_i) = \psi(a_i,c_i^1) \otimes \psi(a_i,c_i^2) \label{eqn:ACPot}\\
&\psi(a_i,c_i^1) = F_{c^1}(a_i,c_i^1)~\mathcal{N}(\|s_{a_i}-s_{c_i^1}\|^2,\mu_{c^1},\sigma_{c^1}) \label{eqn:ACOPot}\\
&\psi(a_i,c_i^2) = \displaystyle \sum_{a\in A} \text{bin}(c_i^2) \mathcal{I}(a=a_i)^T~\mathcal{N}(c_i^2, \mu_{c^2}, \sigma_{c^2}) \label{eqn:ACPPot}
\end{align}

{\bf Activity node potential, $\phi(a_i,x_i)$.} These potentials correspond to the  prior probabilities of the  $a_i$ nodes of the CRF. They describe the inherent characteristics of the activities through low level motion features. We extract low level features $x_i$ from the activity segments $a_i$ and use the pre-trained baseline classifier $\mathcal{P}$ to generate classification scores for these candidate activity segments $a_i$.   We use these scores as the node potential as defined in Equation \ref{eqn:nodeAPot}.

{\bf Context node potential, $\phi(c_i,z_i)$.} These potentials correspond to the prior probabilities of the $V_c$ nodes of the CRF. The context attributes $c_i$ encode the scene-activity context, and are generally scene-level properties and variables representing the presence of specific objects related to the activity of interest. For example, presence of a car may distinguish {\em unloading a vehicle} activity from {\em entering a facility} activity. We compute the context attribute probabilities by applying a number of detectors in the images ($z_i$) of the activity segment ($a_i$) (please see Section \ref{sec:context} for details). The number and type of the context attributes may vary for different applications. For example, we use two context attributes in an application - objects ($\phi(c^1_i,z_i)$) and person ($\phi(c^2_i,z_i)$) attributes as defined in Equations \ref{eqn:objectAttribute} and \ref{eqn:peopleAttribute}, where $c_i^1$ is the object class vector, $c_i^2=\|L_1-L_2\|$ is the distance covered by a person in the activity region, $\text{bin}(\cdot)$ is a binning function as in \cite{R06}, and $\mu_{c^2}$ and $\sigma_{c^2}$ are the mean and variance of the covered distances. We concatenate them in order to compute the context nodes potential (Equation \ref{eqn:contextPot} - $\odot$ is the concatenation operation).

{\bf Activity-Activity edge potential, $\psi(a_i,a_j)$.} This potential models the connectivity among the activities in $A$. We assume that activities which are within a spatio-temporal distance are related to each other. This potential has three components - association, spatial, and temporal components. The association component is the co-occurrence frequencies of the activities. The spatial (temporal) component models the probability of an activity belonging to a particular category given its spatial (temporal) distance from its neighbors. $\psi(a_i,a_j)$ is defined in Equation \ref{eqn:edgeAAPot}, where $a_i,a_j\in V_a$, $F_a(a_i,a_j)$ is the co-occurrence frequency between the activities $a_i$ and $a_j$, $s_{a_i}, s_{a_j}, t_{a_i}, \text{and } t_{a_j}$ are the spatial and temporal locations of the activities, and $\mu_t,\sigma_t, \mu_s, \text{and }\sigma_s$ are the parameters of the Gaussian distribution of relative spatial and temporal positions of the activities, given their categories. 

{\bf Activity-Context edge potential, $\psi(a_i,c_i)$.} This potential function models the relationship among the activities and the context attributes. It corresponds to $V_a-V_c$ edges in the CRF. This potential is defined in Equation \ref{eqn:ACPot}-\ref{eqn:ACPPot}. $\psi(a_i,c_i^1)$ models the relationship between the activity and the object attribute and $\psi(a_i,c_i^2)$ models the relationship between the activity and the person attribute. Operator $\otimes$ performs horizontal concatenation of matrices.

{ \bf Structure Learning.}
The main problem in structure learning is to estimate which activity nodes ($V_a$ nodes) are connected to each other. Note that we do not need to learn the $V_a-V_c$ relationships, because these are established whenever an object is detected by the detector $\mathcal{D}$ in the video segment being considered. However, we need learn the $V_a-V_a$ connections in an online manner because we do not know a priori how the activities are related to each other. A recent approach for learning the structure is hill climbing structure search \cite{YF10}, which is not designed for continuous learning. In this work, we utilize an adaptive threshold based approach in order to determine the connections among the nodes in $V_a$. At first, we assume all the nodes in $V_a$ are connected to each other. Then we apply two thresholds - spatial and temporal - on the links. We keep the links whose spatial and temporal distances are below these thresholds, otherwise we delete the links. We learn these two thresholds using a max-margin learning framework.

Suppose, we have a set of training activities $\{ (a_i, t_{a_i}, s_{a_i}): i  = 1 \ldots m\}$ and we know the pairwise relatedness of these activities from the training activity sequences. We observe which ones in the labeled data happen within a spatio-temporal window and then learn the parameters of that window. The goal is to learn a function $f_r(d) = w^Td$, that satisfies the constraints in Equation \ref{eqn:maxmergin}, where $d_{ij} = [\text{abs}(t_i-t_j), \|s_i-s_j\|]$. 
\begin{align}
f_r(d_{ij}) &= +1, \qquad \forall \text{ related } a_i \text{ and } a_j, \label{eqn:maxmergin}\\
f_r(d_{ij}) &= -1, \qquad \text{otherwise.} \nonumber
\end{align}
We can formulate this problem as a traditional max-margin learning problem \cite{YF10}. Solution to this problem will provide us a function to determine the existence of link between two unknown activities.

{\bf Inference.}
In order to compute the posterior probabilities of the $V_a$ nodes, we choose the belief propagation (BP) message passing algorithm. BP does not provide guarantees of convergence to the true marginals for a graph with loops, but it has proven to have excellent empirical performance \cite{li2008key}. Its local message passing is consistent with the contextual relationship we model among the nodes. At each iteration, the beliefs of the nodes are updated based on the messages received from their neighbors. Consider a node $v_i \in \{V_a, V_c\}$ with a neighborhood $N(v_i)$. The message sent by $v_i$ to its neighbors can be written as, $m_{v_i,v_j}(v_j) = \alpha\int_{v_i}\psi(v_i,v_j)\phi(v_i,x_i) \prod_{v_k\in N(v_i)}m_{v_k,v_i}(v_i)dv_i.$ The marginal distribution of each node $v_i$ is estimated as $\mathcal{P_G}(v_i) = \alpha \phi(v_i,x_i) \prod_{v_j \in N(v_i)} m_{v_j,v_i}(v_i).$ The class label with the highest marginal probability is the predicted class label. We use the publicly available tool \cite{schmidt2012ugm} to compute the parameters of the CRF and to perform the inference.

\section{Context-Aware Instance Selection}
\label{sec:active}
In this section, we describe our method for selecting, from a set of unlabeled activity instances, the most informative ones for manual labeling, so as to improve our recognition models.
Consider that, given a set of past labeled data instances we have learned a baseline classifier $\mathcal{P}$ and a context model $\mathcal{C}$. Now, we receive from the video stream a set of unlabeled activity instances $\mathcal{U} = \{a_i | i = 1, \ldots, N\}$.
We construct a CRF  $G = (V,E)$ with the activities in $\mathcal{U}$ using $\mathcal{P}$ and $\mathcal{C}$ as discussed in Section \ref{sec:modeling}. %
We denote by $V_a = \{a_1, \dots, a_N\}$ the activity nodes in the CRF, and by $E_a=\{(a_i, a_j)~|~a_i ~ \text{and} ~ a_j ~ \text{are linked}\}$ the set of $V_a-V_a$ edges of the CRF. Moreover, we denote the sub-graph containing the activity nodes and their connections by $G_a = (V_a,E_a)$.
Inference on $G$ provides us with (i) the marginal posterior pmf, $\mathcal{P_G}(a_i)$, for each of the activity nodes $a_i$, and (ii) the marginal joint pmf of each pair of nodes connected by an edge, $\mathcal{P_G}(a_i,a_j)$, $(a_i,a_j) \in E_a$.

Our goal is to use the data in $\mathcal{U}$ to improve the model $\mathcal{P}$ and $\mathcal{C}$ with least amount of manual labeling. We achieve this by selecting for manual labeling a subset of nodes in $V_a$,  such that the joint entropy of {\rm all} the nodes, $H(V_a)$,  will be reduced maximally. In what follows, we describe how the joint entropy of all nodes can be (approximately) computed in a computationally efficient manner (Section~\ref{subsec: entropy}), as well as the formulation of the objective function (Section~\ref{sec:obj-func-deriv}) and a novel exact solution for it (Section~\ref{sec:opt-obj-func}). Section~\ref{sec:updates} describes how the new information is employed for incrementally updating the recognition models.

\subsection{Joint Entropy of Activity Nodes}
\label{subsec: entropy}
The joint entropy of the nodes in $V_a$ can be expressed as:
\begin{equation}
\mathcal{H}(V_a)=\mathcal{H}(a_1)+\mathcal{H}(a_2|a_1)+ \dots+\mathcal{H}(a_N|a_1,\dots,a_{N-1})
\label{eqn:HBreak}
\end{equation}
Using the property $\mathcal{I}(a_1,\dots,a_{n-1};a_n)=H(a_n)-H(a_n|a_1,\dots,a_{n-1})$, Eqn. \ref{eqn:HBreak} can be expressed as:
\begin{equation}
\mathcal{H}(V_a)=H(a_1)+\sum_{i=2}^{N}\Big[\mathcal{H}(a_i)-\mathcal{I}(a_1,\dots,a_{i-1};a_i)\Big].
\label{eqn:HBreak1}
\end{equation}
where $\mathcal{I}(\cdot)$ represents the mutual information. Using the chain rule of mutual information (i.e., $\mathcal{I}(a_1,\dots,a_{i-1};a_i) = \sum_{j=1}^{i}\mathcal{I}(a_j;a_i|a_1,\dots,a_{j-1})$), Eqn. \ref{eqn:HBreak1} can be expressed as:
\begin{align}
\mathcal{H}(V_a)&=\mathcal{H}(a_1)+\sum_{i=2}^{N}\Big[\mathcal{H}(a_i)-\sum_{j=1}^{i}\mathcal{I}(a_j;a_i|a_1,\dots,a_{j-1})\Big] \nonumber \\
&= \sum_{i=1}^{N}\mathcal{H}(a_i) - \sum_{i=2}^{N}\sum_{j=1}^{i}\mathcal{I}(a_j;a_i|a_1,\dots,a_{j-1})
\label{eqn:HBreak2}
\end{align}
Computing the conditional mutual information $\mathcal{I}(a_j;a_i|a_1,\dots,a_{j-1})$ is computationally intractable as the number of nodes increases. Moreover, we construct our CRF as a collection of unary (node) and pair-wise (edge) potentials instead of factor or clique graphs. Thus, we can easily approximate the conditional mutual information as $\mathcal{I}(a_j;a_i|a_1,\dots,a_{j-1}) \approx \mathcal{I}(a_j;a_i)$. As a simplifying approximation, here consider two nodes to be independent if there exists no link between them. This allows us to use the property that if two random variables are independent, the mutual information between them is zero. Using these assumptions, Eqn. \ref{eqn:HBreak2} can be written as
\begin{equation}
\mathcal{H}(V_a) =\sum_{i \in V_a}\mathcal{H}(a_i) - \sum_{(i,j) \in E_a}\mathcal{I}(a_j;a_i)
\label{eqn:HBreak3}
\end{equation}
We use this expression to derive an objective function to be optimized in order to obtain the most informative activity nodes for manual labeling.

\subsection{Objective Function Derivation}
\label{sec:obj-func-deriv}

\begin{figure}
\centering
\includegraphics[scale=0.4]{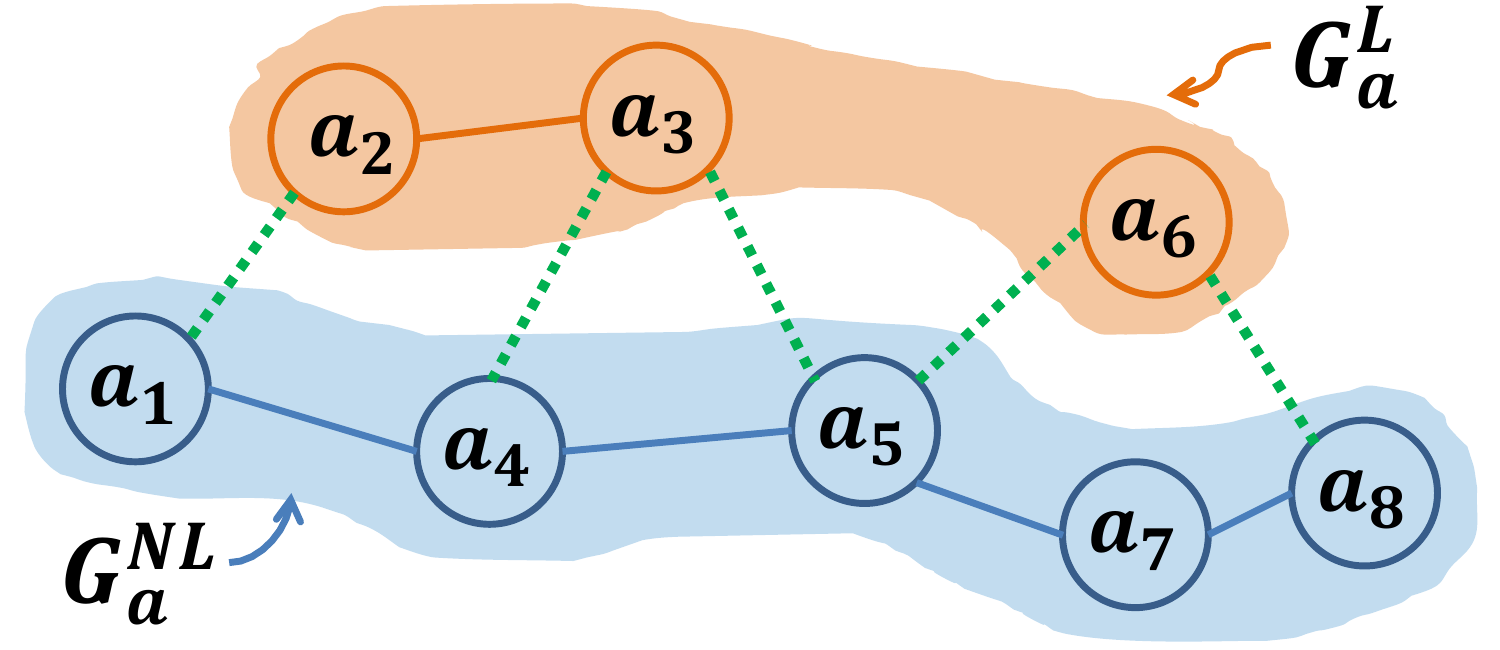}
\caption{This figure illustrates the partitioning of the graph $G_a$ of activity nodes into two subgraphs, (i) the graph $G_a^L$, whose nodes should be queried for labeling and (ii) the graph $G_a^{NL}$ which is not labeled. The green dotted lines denote the links between the two subgraphs. As in Eqn. \ref{eqn:HBreak4}, the joint entropy (J.E.) of the entire graph can be expressed as summation of the J.E. of two the subgraphs minus the mutual information of the links in between them.  }
\label{subgraph}
\end{figure}

Let us consider that we select a subset of $K$ activity instances for manual labeling  from the set $\mathcal{U}$ ($K \le N$ and depends on manual labeling budget). Since these activity instances are nodes of graph $G_a$, a subgraph can be formed by using these nodes. Consider $G^L_a=(V^L_a,E^L_a)$ to be a subgraph of the graph $G_a$, where $V^L_a$ are the $K$ nodes chosen for manual labeling and $E^L_a=\{(a_i,_j)|(a_i,a_j)\in E_a, a_i, a_j \in V^L_a\}$. Then the remaining nodes which are not selected for manual labeling also constitute a subgraph $G^{NL}_a=(V^{NL}_a,E^{NL}_a)$, where $V^{NL}_a = V_a - V^L_a$ and $E^{NL}_a = \{(a_i,a_j)|(a_i,a_j) \in E_a, a_i, a_j \in V^{NL}_a\}$. This partitioning is presented pictorially in Fig. \ref{subgraph}. The joint entropy $H(V_a)$ can be partitioned as follows:
\begin{align}
\mathcal{H}(V_a) &= \Big[\sum_{i \in V^L_a}\mathcal{H}(a_i)-\sum_{(i,j) \in E^L_a}\mathcal{I}(a_j;a_i)\Big] +  \Big[\sum_{i \in V^{NL}_a}\mathcal{H}(a_i) \nonumber \\
&- \sum_{(i,j) \in E^{NL}_a}\mathcal{I}(a_j;a_i)\Big] - \Big[\sum_{\substack{(i,j) \in E_a \\i \in V^L_a, j \in V^{NL}_a}} \mathcal{I}(a_j;a_i)\Big] \nonumber \\
& =\mathcal{H}(V^L_a) + \mathcal{H}(V^{NL}_a) - \sum_{\substack{(i,j) \in E_a \\i \in V^L_a, j \in V^{NL}_a}} \mathcal{I}(a_j;a_i)
\label{eqn:HBreak4}
\end{align}
The first and the last terms in the above equation will be zero if the nodes in $V^L_a$ are manually labeled (please see the proof in Appendix A). Since our goal is to choose $K$ nodes from $V_a$ for manual labeling such that the joint entropy $\mathcal{H}(V_a)$ decreases maximally, the optimal subset of nodes to be chosen for manual labeling can be expressed as:
\begin{equation}
V^{L*}_a = \argmax_{\substack{V^L_a \\ s.t. |V^L_a|=K}} \Big[ \mathcal{H}(V^L_a) - \sum_{\substack{(i,j) \in E_a \\i \in V^L_a, j \in V^{NL}_a}} \mathcal{I}(a_j;a_i) \Big]
\label{eqn:HBreak5}
\end{equation}
The above function can be simplified as follows:

\begin{align}
&\mathcal{F}(V^L_a) = \mathcal{H}(V^L_a) - \sum_{\substack{(i,j) \in E_a \\i \in V^L_a, j \in V^{NL}_a}} \mathcal{I}(a_j;a_i) \nonumber \\
&= \sum_{i \in V^L_a}\mathcal{H}(a_i) - \sum_{\substack{(i,j) \in E_a \\ (i,j) \notin E^{NL}_a}} \mathcal{I}(a_j;a_i) \nonumber \\
&= \sum_{i \in V^L_a}\mathcal{H}(a_i) - \Big[ \sum_{(i,j) \in E_a} \mathcal{I}(a_j;a_i) -\sum_{(i,j) \in E^{NL}_a}\mathcal{I}(a_j;a_i)\Big]
\label{eqn:HBreak6}
\end{align}
We need to choose nodes from $V_a$ to be in $V^L_a$ for labeling, such that the above expression is maximized. Consider a vector $\boldsymbol{u}$ of length $N$ with elements either $1$ or $0$, where a $1$ in the $i$-th position represents that the corresponding node has been selected for $V^L_a$ and $0$ represents the opposite. Thus, we need to find the optimal $\boldsymbol{u}$ such that $\mathcal{F}(V^L_a)$ is maximized. In order to rewrite the objective function into a convenient matrix format, let us define a $N \times 1$ vector $\boldsymbol{h}$ of node entropies and a $N \times N$ matrix $\boldsymbol{M}$ of pairwise mutual information as follows:
\begin{align*}
\boldsymbol{h} \triangleq [\mathcal{H}(a_1), \mathcal{H}(a_2) \dots \mathcal{H}(a_N)]^T \\
\boldsymbol{M}(i,j) \triangleq
\begin{cases}
    \mathcal{I}(a_i;a_j),& \text{if } (i,j) \in E_a \\
    0,                   & \text{otherwise}
\end{cases}
\end{align*}
\noindent
where $i,j \in \{1,\dots,N\}$. With this notation, the objective function in Eqn. \ref{eqn:HBreak6} can be represented as a function of $\boldsymbol{u}, \boldsymbol{h}$ and $\boldsymbol{M}$ as follows:
\begin{align}
\mathcal{G}(\boldsymbol{u}) &= \boldsymbol{u}^T\boldsymbol{h} - \dfrac{1}{2}\Big[ \boldsymbol{1}^T\boldsymbol{M}\boldsymbol{1}-(\boldsymbol{1}-\boldsymbol{u})^T\boldsymbol{M}(\boldsymbol{1}-\boldsymbol{u}) \Big] \nonumber \\
&= \boldsymbol{u}^T\boldsymbol{h} - \boldsymbol{u}^T\boldsymbol{M}\boldsymbol{1} + \dfrac{1}{2}\boldsymbol{u}^T\boldsymbol{M}\boldsymbol{u}
\label{Gx}
\end{align}
\noindent
where $\boldsymbol{1}$ is an $N\times 1$ vector of ones. Maximizing $\mathcal{G}(\boldsymbol u)$ is equivalent to minimizing $-\mathcal{G}(\boldsymbol u)$. Therefore, the optimization problem in \ref{eqn:HBreak5} can be reformulated as
\begin{align}
&\boldsymbol{u^*} = \argmin_{\boldsymbol{u}} \dfrac{1}{2} \boldsymbol{u}^T \boldsymbol{Q} \boldsymbol{u} + \boldsymbol{u}^T\boldsymbol{f} \nonumber \\
& s.t. \quad \boldsymbol{u}^T\boldsymbol{1} = K, \quad \boldsymbol{u} \in \{0,1\}^N
\label{opt1}
\end{align}
where $\boldsymbol{Q} \triangleq -\boldsymbol{M}$ and $f \triangleq  \boldsymbol{M} \boldsymbol{1}-\boldsymbol{h}$. The procedure followed to solve the above optimization problem is discussed next.

\subsection{Optimization of Objective Function}
\label{sec:opt-obj-func}
The matrix $\boldsymbol{Q}$ in Eqn. \ref{opt1} is not positive semi-definite (please refer to Appendix B for details), thus the objective function is non-convex. The second constraint in this optimization problem is also non-convex. Thus, Eqn. \ref{opt1} is a non-convex binary quadratic optimization problem. However, due to the binary constraints on $\boldsymbol{u}$, a constant diagonal matrix $\gamma \boldsymbol{I}$ can be added to the objective function to make it convex, where $\boldsymbol{I}$ is an identity matrix of size $N \times N$ and $\gamma \geq \max \{ |\boldsymbol{M}|\boldsymbol{1}\}$ is a constant (please refer to Appendix C for details). This is because adding $\gamma \boldsymbol{I}$ to the objective function is equivalent to adding a constant $K\gamma$ at all feasible points of the optimization problem in Eqn. \ref{opt1}.  Thus, the optimization problem in in Eqn. \ref{opt1} is equivalent to the following one:
\begin{align}
&\boldsymbol{u^*} = \argmin_{\boldsymbol{u}} \dfrac{1}{2} \boldsymbol{u}^T (\boldsymbol{Q}+\gamma \boldsymbol{I}) \boldsymbol{u} + \boldsymbol{u}^T\boldsymbol{f} \nonumber \\
& s.t. \quad \boldsymbol{u}^T\boldsymbol{1} = K, \quad \boldsymbol{u} \in \{0,1\}^N
\label{opt2}
\end{align}
The above objective function is convex, but the second constraint remains non-convex. We use the branch and bound (BB) method \cite{boyd2007branch} to solve this problem. At each node of BB, we relax the second constraint as $0 \leq u_i \leq 1$ (where $u_i$ denotes the $i^{th}$ element of $\boldsymbol{u}$) and solve the relaxed convex optimization problem using the CVX Solver \cite{grant2008cvx}. Importantly, the resulting solution is guaranteed to be globally optimal using this method. Although in the worst case BB can end up solving $\binom{N}{K}$ convex problems, on average a much smaller number of convex problems needs to be solved before reaching the optimal solution. In fact, for all the experiments executed in this paper, BB has reached the globally optimal solution in a significantly smaller  amount of time (approximately fraction of a second) than the worst-case prediction.

\begin{figure}[ht]
\centering
\includegraphics[scale=0.6]{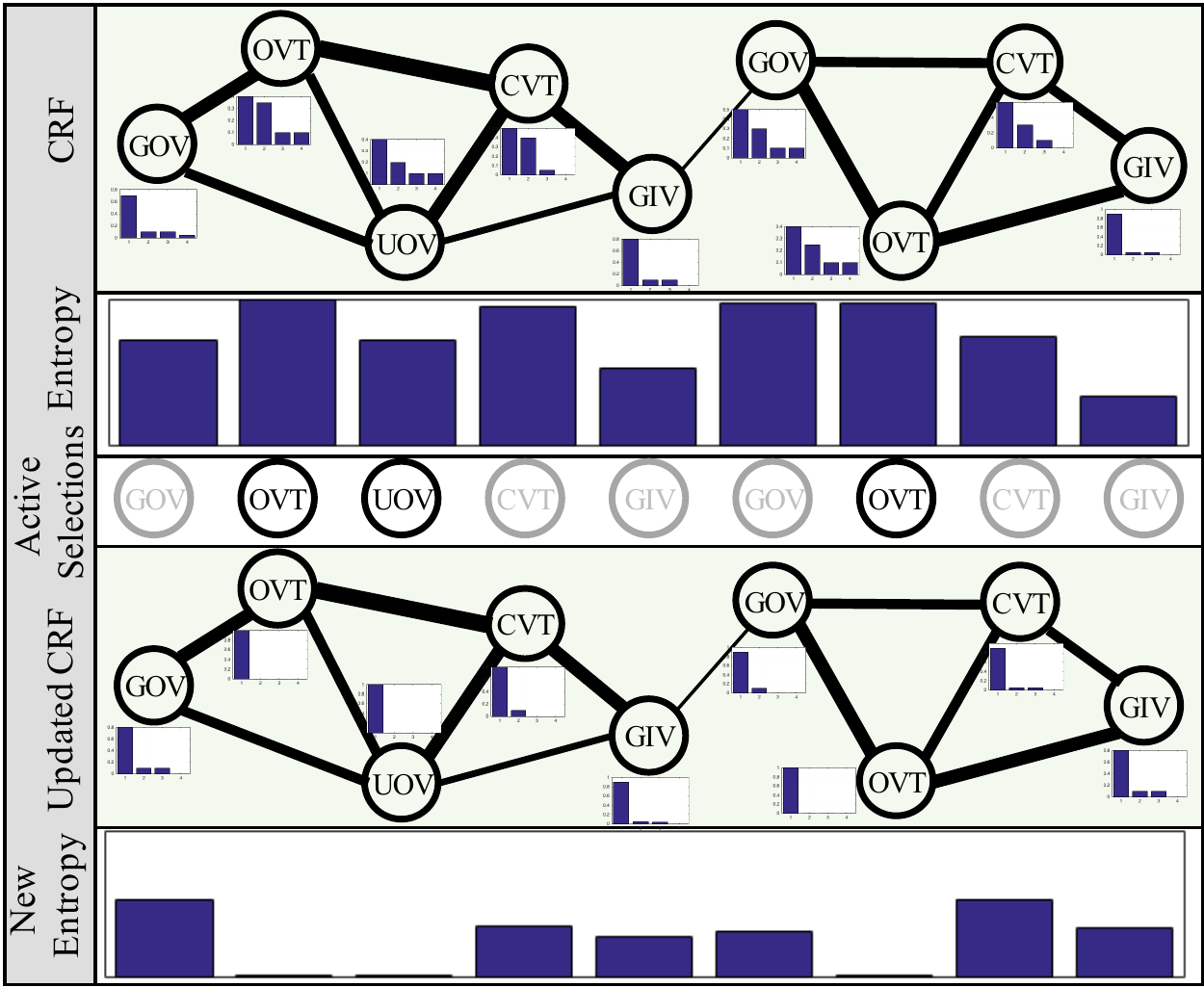}
\caption{An example run of our proposed active learning framework on a part of activity sequence from VIRAT dataset. Circles are activity nodes along with their class probability distribution. Edges have different thickness based on the pairwise mutual information. The node labels are - getting out of the vehicle (GOV), opening vehicle trunk (OVT), unloading from vehicle (UOV), closing vehicle trunk (CVT), and getting into the vehicle (GIV). Inference on the CRF (top) gives us marginal probability distribution of the nodes and edges. We use these distributions to compute entropy and mutual information. Relative mutual information is shown by the thickness of the edges, whereas entropy of the nodes are plotted below the top CRF. Equation \ref {eqn:HBreak3} exploits entropy and mutual information criteria in order to select the most informative nodes (2-OVT, 3-UOV, and 7-OVT). We condition upon these nodes (filled) and perform inference again, which provides us more accurate recognition and a system with lower entropy (bottom plot).}
\label{fig:al-example}
\end{figure}

We ask a human annotator (strong teacher) to label the instances in  $V^{L*}_a$. We then perform inference on $G$ again by conditioning on the nodes $a_i \in V^{L*}_a$. This provides more accurate (and more confident) labels to the remaining nodes in $G$. At this time, for an instance $a_j \in V^{NL}_a$, if one of the classes has probability greater than $\delta$ (say $\delta=0.9$), we assume that the current model $\mathcal{P_G}$ is highly confident about this instance. We retain this instance along with its label obtained from the inference for incremental training. We refer to this as the weak teacher. The number of instances obtained from the weak teacher depends on the value of $\delta$, which we choose to be large for safety, so that miss-classified instances are less likely to be used in incremental training. An illustrative example of our active learning system is shown in Figure \ref{fig:al-example}.

\subsection{Incremental Updates}
\label{sec:updates}

Given the newly available labeled samples, we then proceed to update the activity recognition model and the context models. These models are responsible for the node and edge potentials of the CRF respectively.

{\bf Updating activity recognition model.}
In our experiments, we use two different activity recognition models as the baseline activity classifiers: multinomial logistic regression (MLR) and support vector machine (SVM). For the SVM classifier, we use \cite{poggio2001incremental} for incrementally updating its parameters. We next describe how to incrementally update the MLR model parameters, given the new labeled instances.

In the MLR model, the probability of activity $a_i$ belonging to class $A_j$ is computed as:
\begin{align}
p(a_i \in A_j | x_i; \boldsymbol{\theta}) = \frac{\exp (\boldsymbol{\theta}_j^T x_i)}{\sum_{l=1}^q \exp ({ \boldsymbol{\theta}_l^T} x_i)},
\end{align}
where $\boldsymbol{\theta}_j$, $j = 1,\hdots, q$ is the weight vector corresponding to class $j$. To obtain the optimal weight vectors, we seek to minimize the following cost function:
\begin{align}
\argmin_{\boldsymbol{\theta}}J(\boldsymbol{\theta}) = -\frac{1}{m} \sum_{i=1}^{m} \sum_{j=1}^{q} \mathbf{1} \{a_i \in A_j\} \notag \\ \log  p\left( a_i \in A_j|x_i;\boldsymbol{\theta} \right)
 + \frac{\lambda}{2}  \|\theta\|^2
 \label{eqn:softmax}
 \end{align}
where  $m$ is the number of training instances available for the incremental update, $\mathbf{1}\{.\}$ is the identity function, $\lambda$ is the weight-decay parameter, and $\|.\|$ is the $l_2$ norm. This is a convex optimization problem and we solve it using gradient descent, which provides a globally optimal solution. The gradient with respect to $\boldsymbol \theta_j$ can be written as
\begin{align}
\nabla_{\boldsymbol{\theta_j}} J(\boldsymbol{\theta}) = - \frac{1}{m} \sum_{i=1}^{m}{ \left( x_i \left( \mathbf{1}\{ a_i \in A_j \}  - p(a_i \in A_j| x_i; \boldsymbol{\theta}) \right) \right)}.
\end{align}
For updating the MLR model, we obtain the newly labeled instances from both the strong the the weak teacher and store them in a buffer. When the buffer is full, we use all of these instances to compute the gradients $\nabla_{\boldsymbol{\theta_j}} J(\boldsymbol{\theta})$, $j= 1,\hdots, q$ of the model. Then we update the model parameters using gradient descent as follows:
\begin{align}
\boldsymbol{\theta}_j^{t+1} = \boldsymbol{\theta}_j^t - \alpha \nabla_{\boldsymbol{\theta}_j^t} J(\boldsymbol{\theta}),
\end{align}
where $\alpha$ is the learning rate. This technique is known as the mini-batch training in literature \cite{S02}, where model changes are accumulated over a number of instances before applying updates to the model parameters.

{\bf Updating context model.}
Updating the context model is consists of updating the parameters of the Equations \ref{eqn:peopleAttribute}, \ref{eqn:edgeAAPot}, \ref{eqn:ACOPot}, and \ref{eqn:ACPPot}. The parameters are (i) the co-occurrence frequencies of activities and/or context attributes, and (ii) the means and variances of the Gaussian distributions used in the activity relationship models. The parameters of the Gaussians can be updated using the method in \cite{ross2008incremental}, whereas the co-occurrence frequency matrices $F_a$ and $F_c$ can be updated as follows:
\begin{align}
F_{ij} = F_{ij} + \text{sum}([(L=i).(L=j)^T].*Adj),
\end{align}
where $i\in \{A_1, \ldots, A_q\}$, $j \in \{A_1, \ldots, A_q\}$ (for $F_a$), $j \in c_i^1$ (for $F_c$), $L$ is the set of labels of the instances in $\mathcal{U}$ obtained after the inference, $Adj$ is the adjacency matrix of the CRF $G$ of size $|L| \times |L|$, $\text{sum}(.)$ is the sum of the elements in the matrix, and $.*$ is the element wise matrix multiplication.

The overall framework is portrayed in the supplementary.

\section{Experiments}
\label{sec:expt}
We conduct experiments on six challenging datasets - UCF50 \cite{RS12}, VIRAT \cite{SHP+11}, UCLA-Office \cite{pei2011parsing}, MPII-Cooking \cite{rohrbach2012database}, AVA \cite{AVA}, and 50Salads \cite{stein2013combining} - to evaluate the performance of our proposed active learning framework.  We briefly describe these datasets as follows. Dataset descriptions and experiment details for UCLA-Office can be found in the supplementary. 

{\bf Activity segmentation.}
For VIRAT and UCLA-Office, we use an adaptive background subtraction algorithm to identify motion regions. We detect moving persons around these motion regions using \cite{FGM} and use them to initialize a tracking method in order to obtain local trajectories of the moving persons. We collect STIP features \cite{L05} from these local trajectories and use them as the observation in the method proposed in \cite{CRH+09} to identify candidate activity segments from these motion regions. Activities are already temporally segmented in UCF50, whereas for MPII-Cooking we use the segmentation provided with the dataset.

{\bf Baseline Classifier.} We use multinomial logistic regression or softmax as the baseline classifier for VIRAT, UCLA-Office, and UCF50 datasets, whereas we use linear SVM for the MPII-Cooking dataset. Please note that the choice of our classifier is based on whichever performs the best of each dataset; this does not affect the interpretation of our results as our method is classifier agnostic.

{\bf Appearance and motion features.} We use C3D \cite{tran2014learning} features as a generic feature descriptor for video segments for the UCF50 and VIRAT datasets. C3D exploits 3D convolution that makes it better than conventional 2D convolution for motion description. We use an off-the-self C3D model trained on the Sports-1M \cite{KarpathyCVPR14} dataset. Given the video segment, we extract a C3D feature of size 4096 for each sixteen frames with a temporal stride of eight frames. Then, we max pool the features in order to come up with a fixed-length feature vector for the video segment.

For the UCLA-Office dataset, we extract STIP \cite{L05} features from the activity segments. We use a video feature representation technique based on spatio-temporal pyramid and average pooling similar to \cite{hasan2014Continuous} to compute a uniform representation using these STIP features. 

For the MPII-Cooking dataset, we use a bag-of-words based motion boundary histogram (MBH) \cite{dalal2006human} feature that comes with the dataset. Note that our framework is independent of any particular feature or video representation. It allows us to plug in the best video representation for any application.

\subsection{Context attributes}
\label{sec:context}	

The number of context features and their types may vary based on the datasets. Our generalized CRF formulation can take care of any number and type of context features. We use co-occurrence frequency of the activities and the objects, their relative spatial and temporal distances, movement of the objects and persons in the activity region, etc. as the context feature. Some of the features were described in Section \ref{sec:modeling}. Context features naturally exist in VIRAT, UCLA-Office, and MPII-Cooking datasets. For UCF50, we improvise a context feature by assuming that similar types of activities co-occur in the nearby spatial and temporal vicinity. Dataset specific detailed description of these features are described below.

{\bf VIRAT and UCLA-Office:} We use both of the scene-activity and the inter-activity context features for these datasets. They have been described in Equations \ref{eqn:energy} to \ref{eqn:ACPPot}.	We compute scene-activity context features using object detections, whereas, inter-activity context features are computed using the spatial temporal relationships such as co-occurrence frequency among the activities.

{\bf MPII-Cooking:} Similar to previous two datasets, we use both of the scene-activity and the inter-activity context features for this dataset. While inter-activity context features remains same, scene-activity context is different from previous two datasets. Activities in this dataset involve three types of objects - tools ($c_i^1$), ingredients ($c_i^2$), and containers ($c_i^3$). We use each of them as a separate context and formulate them as in Equations \ref{eqn:peopleAttribute}, \ref{eqn:objectAttribute}, \ref{eqn:ACOPot}, and \ref{eqn:ACPPot}. 3, 4, 7, and 8. So the Equations \ref{eqn:contextPot} and \ref{eqn:ACPot} become, $\phi(c_i,z_i) = \phi(c_i^1,z_i) \odot \phi(c_i^2,z_i) \odot \phi(c_i^3,z_i)$ and $\psi(a_i,c_i) = \psi(a_i,c_i^1) \otimes \psi(a_i,c_i^2) \otimes \psi(a_i,c_i^3)$.

\def \ss{0.26}
\begin{figure*}[h]
	\centering
	\begin{tabular}{ccc}
		\includegraphics[scale=\ss]{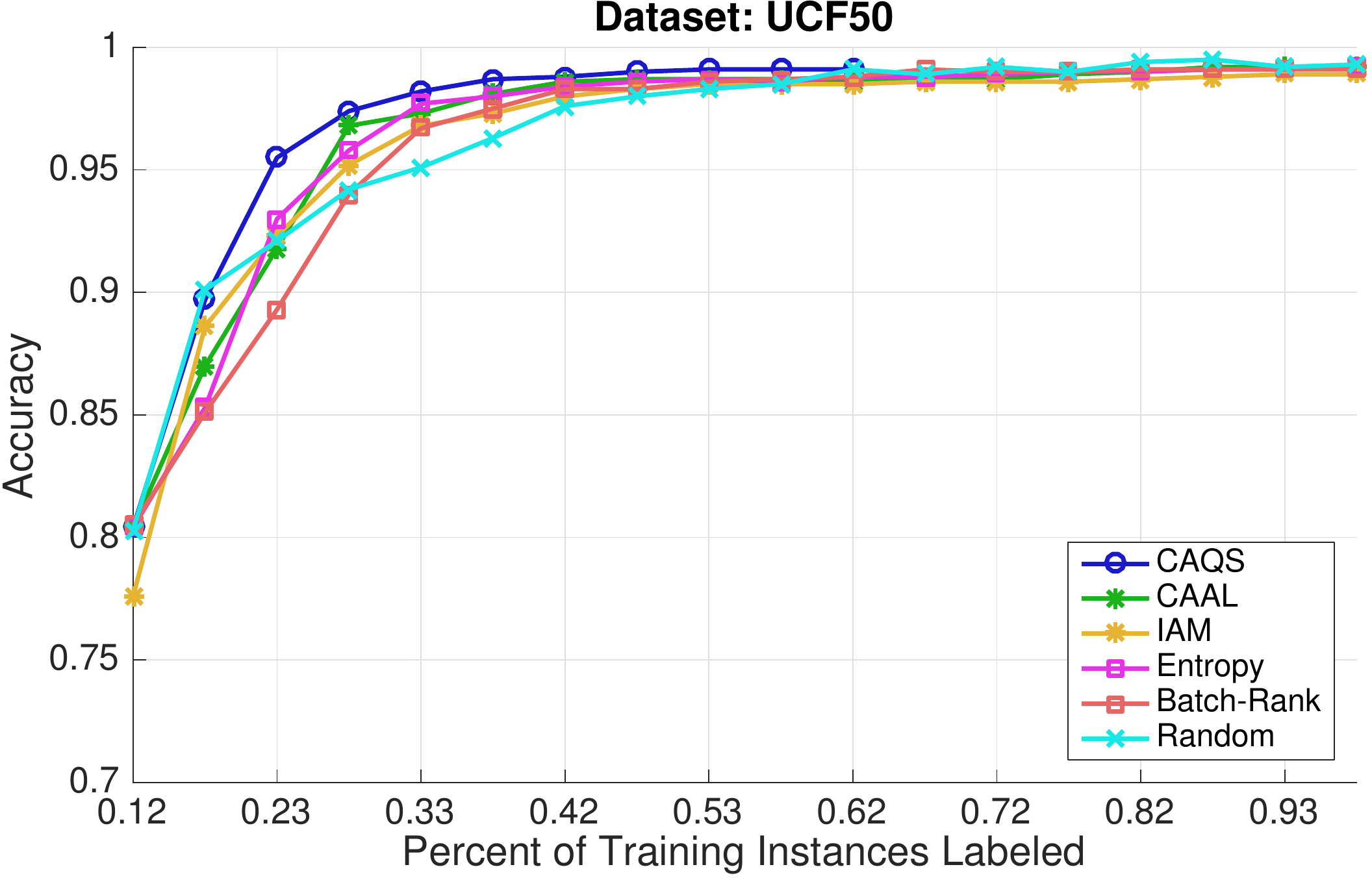} &
		\includegraphics[scale=\ss]{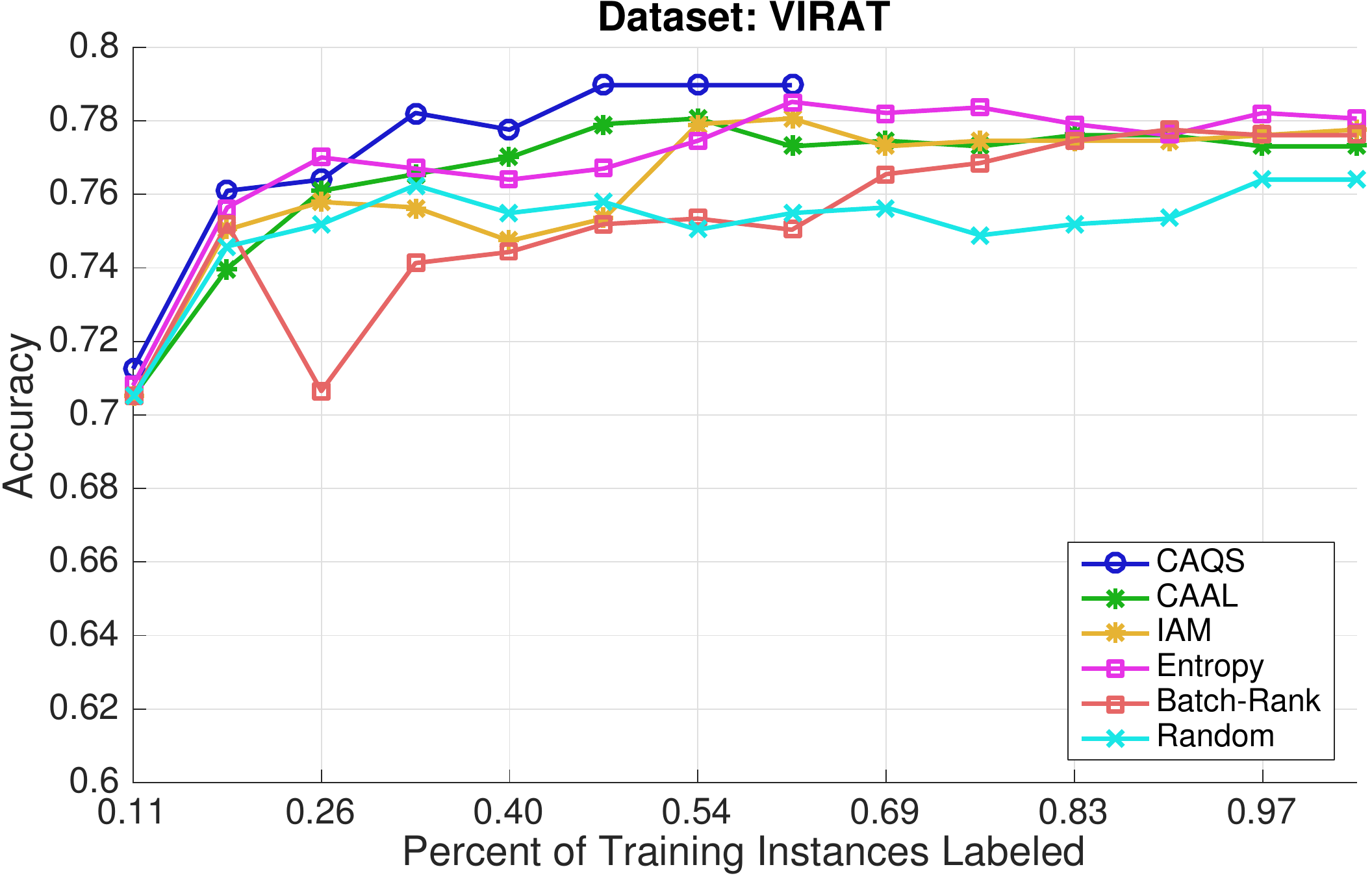} &
		\includegraphics[scale=\ss]{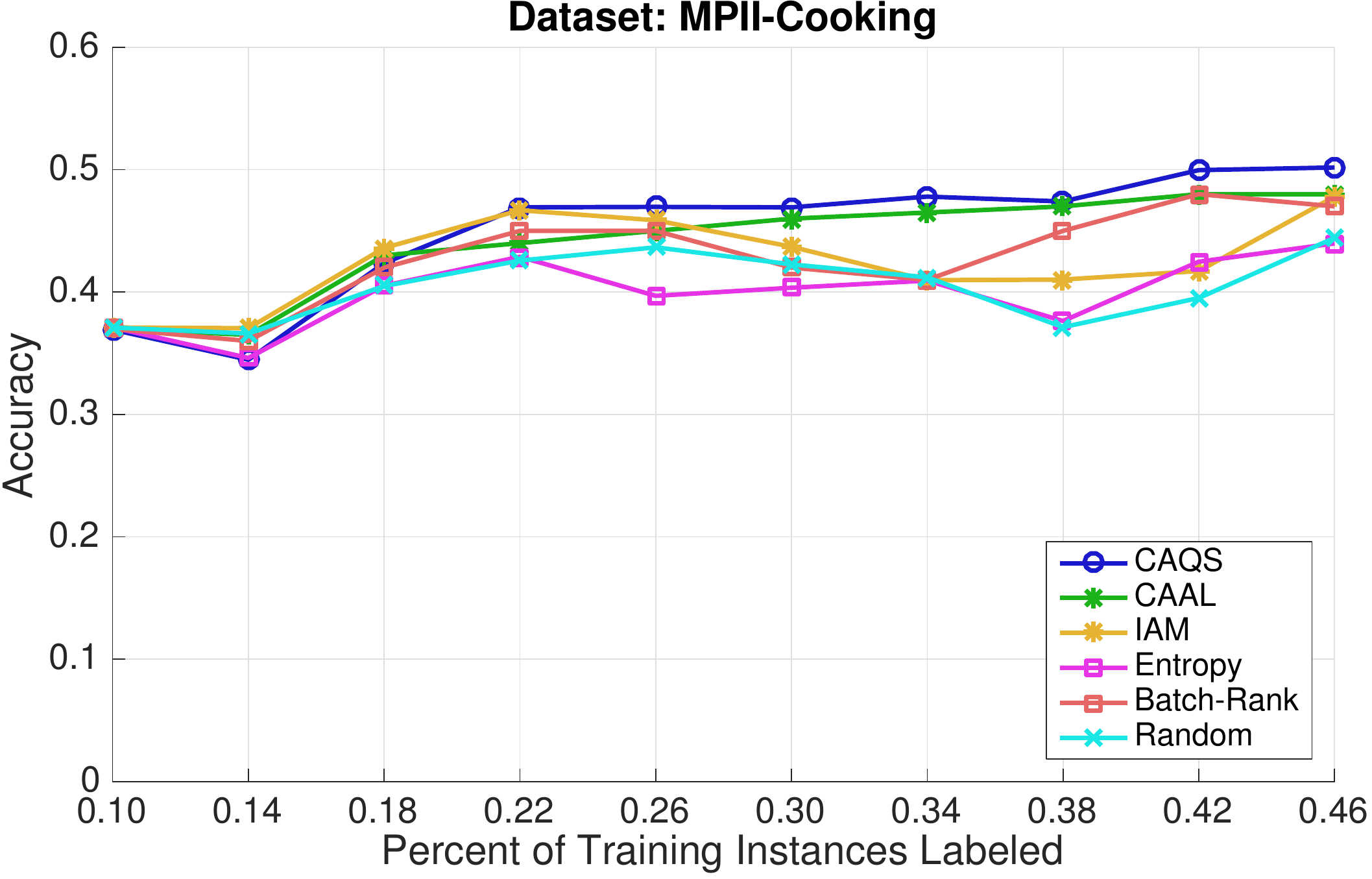} \\
		(a) & (b) & (c)
	\end{tabular}
	\caption{Performance comparison against other competitive active learning methods on four datasets such as (a) UCF50, (b) VIRAT, and (c) MPII-Cooking. The X-axis represents the number of manually labeled training instances, whereas the Y-axis represents correct recognition accuracy on a set of unseen test instances. Please see the text in Section \ref{expt:comp_al} for detailed explanation. Best view in color.}
	\label{plot:comps_al}
\end{figure*}

{\bf UCF50:} Since the activities in UCF50 dataset are provided as individual segments, there are no natural spatial-temporal relationships that exist among them. Also, each of the activities involves a person and one particular tool, so the use of object context might overfit the model, as no context-sharing exists between activities. Thus, similar to \cite{hasan2015context} we use a relationship among the activities based on the activity super-categories and the likelihood of them happening together. We categorize fifty activity classes into eight super-categories, where activities are inter-related. We arrange the individual activities into sequences, where the activities belong to the same super-category are placed nearby. Therefore, they enhance the recognition of each other during inference. These super-categories are provided in the supplementary. The corresponding mathematical formulations remain the same as in Equations \ref{eqn:nodeAPot} and \ref{eqn:edgeAAPot}.

\subsection{Experiment Setup}
\label{subsec:Exp setup}

\begin{figure*}[h]
 	\centering
 	\begin{tabular}{ccc}
 		\includegraphics[scale=\ss]{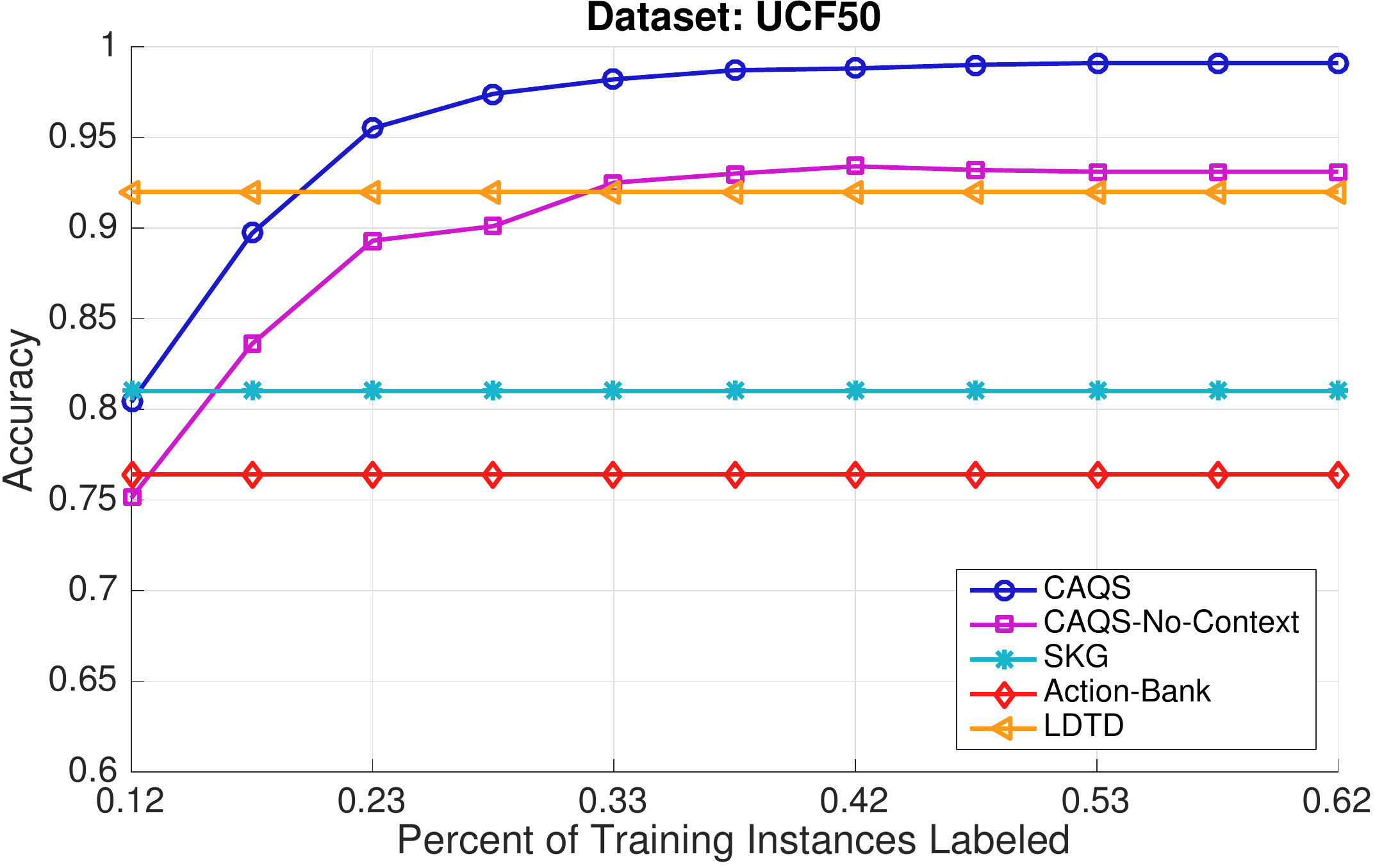}&
 		\includegraphics[scale=\ss]{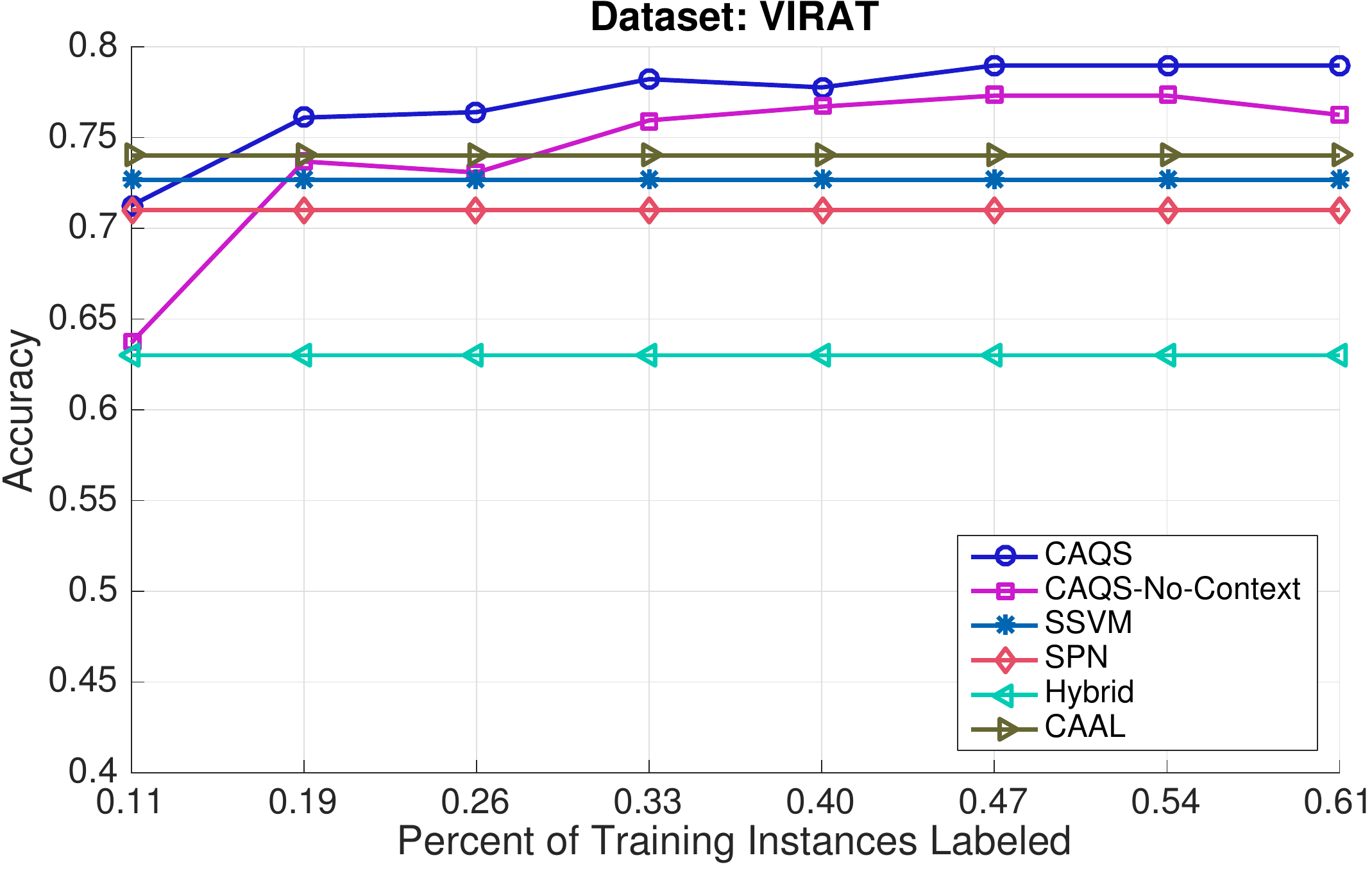} &
 		\includegraphics[scale=\ss]{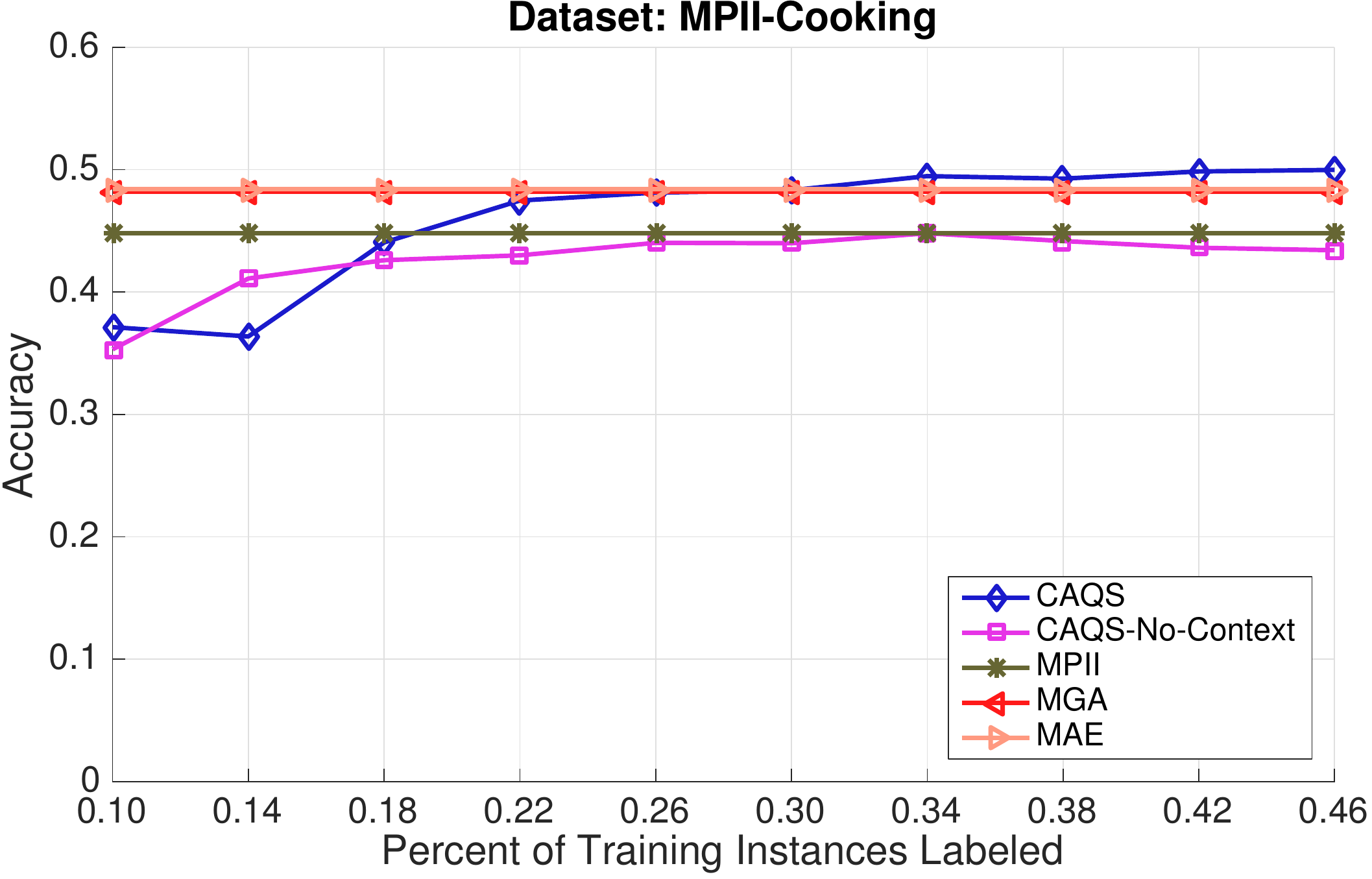} \\
 		(a) & (b)  & (c)\\
 	\end{tabular}
 	\caption{Performance comparison against other state-of-the-art batch and incremental methods. The X-axis represents the number of manually labeled training instances, whereas the Y-axis represents correct recognition accuracy on a set of unseen test instances. Please see the text in Section \ref{expt:comp_others} for detailed explanation. Best view in color.}
 	\label{plot:comps_others}
 \end{figure*}

The training data is sequential in time, where activities occurring within a temporal vicinity are inter-related. For datasets like VIRAT, UCLA-Office, and MPII-Cooking, where video sequences are long and contain multiple activities, these interrelationships are natural. However, for UCF50, we enforce this temporal relationship by using the fact that similar types of activities tend to co-occur as mentioned earlier. Given the training data in a sequence, we use approximately ten percent of them in the initial training phase to train the initial recognition and context models. This initial batch of data is manually labeled. We iteratively update these models using rest of the training data. At each iteration, we select the $K$ most informative instances and use them to update both of the recognition and the context models. Additionally, at each iteration, we evaluate the performance of updated models on the unseen test data and report the accuracy. The reported accuracies are computed by dividing the number of correct recognitions by the number of instances presented. We evaluate the experimental results as follows:

\begin{itemize}[leftmargin=*]
\item Comparison of the proposed Context Aware Query Selection (CAQS) method against other state-of-the-art active-learning techniques (Figure \ref{plot:comps_al}).
\item Comparison with other batch and incremental methods against two different variants of our approach based on the use of context attributes such as CAQS and CAQS-No-Context (Figure \ref{plot:comps_others}). While CAQS utilizes context attributes along with active and incremental learning, CAQS-No-Context does not exploit any explicit context attributes. 
\item Performance evaluation of the four variants of our proposed active-learning framework based on the use of strong and weak teachers. (Figure \ref{plot:comps_self}).
\end{itemize}

\subsection{Comparison with Active-Learning Methods}
\label{expt:comp_al}

Plots in Figure \ref{plot:comps_al} illustrate the comparisons of our context-aware query selection for active learning (CAQS) method against random sampling and four other state-of-the-art active learning techniques: CAAL \cite{hasan2015context}, IAM \cite{HC14}, Entropy \cite{druck2009active}, and Batch-Rank \cite{Chakraborty2015Active}. CAAL exploits both  the entropy and the mutual information in order to select the most informative queries but only provides a greedy solution for query selection. IAM selects a query by utilizing the classifier's decision ambiguity over an unlabeled instance and takes advantage of both weak and strong teachers. It measures the difference between the top two probable classes. If the difference is below a certain threshold, the instance is selected for manual labeling. Entropy \cite{druck2009active} selects a query if the classifier is highly uncertain about it based on the entropy measure. Batch-Rank solves a convex optimization problem that contains entropy and KL-divergence in order to select the instances to be labeled by a human.  We follow same experiment setup  and parameters for these experiments for ensuring fairness.

The plots show that proposed CAQS outperforms other active-learning techniques and random sampling over all datasets. This is because our method can efficiently utilize the interrelationships of the instances using a CRF. Additional observations regarding the performance are as follows:

\begin{itemize}[leftmargin=*]
\item All the plots eventually saturate toward a certain accuracy after some amount of manual labeling. This is because by that point the methods have already learned most of the information present in the training data. The rest of the instances possess little information with respect to the current model.

\item CAQS reaches the saturation-level accuracy the quickest among all methods tested. Its accuracy sharply increases with the amount of manual labeling. This is because it can efficiently select the most informative training instances and learn the best classifier that results in higher recognition accuracy with less amount of manual labeling.

\item The performance of the random sampling is the worst, as expected, and the performance of the other methods is between CAQS and random sampling.

\item Even though CAQS performed better than CAAL, the margins are not significant. This is because both use a similar optimization criterion for query selection. CAAL provides a greedy solution, wheres CAQS provides a solution with global optimality guarantees.
\end{itemize}

\subsection{Comparison with State-of-the-Art Methods}
\label{expt:comp_others}

The plots in Figure \ref{plot:comps_others} illustrate the comparison of our two test cases - CAQS and CAQS-No-Context against state-of-the-art batch and incremental methods on four datasets. The definitions of these two test cases are as follows. CAQS-No-Context means we apply the activity recognition model $\mathcal{P}$ independently on the activity segments without exploiting any spatio-temporal contextual information. CAQS context means we exploit the object and person attribute context along with the $V_a-V_a$ context (Fig. \ref{fig:crf}). In both of these two cases, we use active learning with both of the weak and the strong teachers.

We compare the results on UCF50 datasets against stochastic Kronecker graphs (SKG) \cite{todorovic2012human}, action bank \cite{SC12}, and learned deep trajectory descriptor (LDTD) \cite{shi2015learning}. We compare the results on the VIRAT dataset against structural SVM (SSVM) \cite{ZNR13}, sum product network (SPN) \cite{AT12}, Hybrid \cite{hasan2014Continuous}, and CAAL \cite{hasan2015context}. We compare the results on MPII-Cooking dataset against MPII \cite{rohrbach2012database}, multiple granularity analysis (MGA) \cite{ni2014multiple}, and mid-level action elements (MAE) \cite{lan2015action}. Since these are the batch methods, we report only the final performance of these methods, using all the training instances. Hence, plots of accuracies of these methods are horizontal straight lines. We compared our work with two structure learning methods such as SSVM and SCSG. SSVM learns the structure with structural SVM and SCSG learns the structure with AND-OR graphs. We can observe the following:
\begin{itemize}[leftmargin=*]
\item All of these different datasets show similar asymptotic characteristics. Performance improves with newly labeled training instances.
\item Performance improves when we use contextual information.  CAQS performs better than CAQS-No-Context.
\item Our methods outperform other state-of-the-art batch and incremental methods, using far lower amount of manually labeled data. In these plots our method uses 30\%-40\% manually labeled data depending on the dataset, whereas all other methods use all the instances to train their models. Even though SCSG performs better than CAQS by $1.7\%$, CAQS consumes only $33\%$ manually labeled data compared to $100\%$ of SCSG.
\end{itemize}
Table \ref{table:compare_other} summarizes the performance comparison against other state-of-the-art methods.

\begin{table}[h]
	\centering
	\begin{tabular}{|r|r|p{1cm}|r|p{1cm}|}
		\hline
		& \multicolumn{2}{|c|}{Our Methods} & \multicolumn{2}{|c|}{State-of-the-art}\\
		\hline
		Datasets & Accuracy(\%)& Manual-Labeling & Accuracy(\%) & Manual-Labeling\\
		\hline
		UCF50 &CAQS: $98.2$ & $38\%$&AB: $76.4$& $100\%$\\
		 &CAQS-NoC: $92.5$ & $38\%$&SKG: $81.0$ & $100\%$\\
		 &CAQS: $99.1$ & $100\%$ &LDTD: $92.0$ & $100\%$\\
		 &CAQS-NoC: $93.1$ & $100\%$ &CAAL: $68.0$ & $52\%$\\
		\hline
		VIRAT &CAQS: $77.2$ & $33\%$ &SSVM: $73.5$ & $100\%$\\
		&CAQS-NoC: $75.9$ & $47\%$ &SPN: $71.0$ & $100\%$\\
		&CAQS: $78.9$ & $100\%$ &CAAL: $74.0$ & $42\%$\\
		&CAQS-NoC: $76.3$ & $100\%$ & &\\
		\hline
		MPII &CAQS: $49.4$ & $42\%$ &MPII: $44.8$ & $100\%$\\
		&CAQS-NoC: $44.8$ & $42\%$&MGA: $48.2$ & $100\%$\\
		&CAQS: $49.6$ & $100\%$ &MAE: $48.4$ & $100\%$\\
		&CAQS-NoC: $45.2$ & $100\%$ &CAAL: $48.5$ & $44\%$\\
		\hline
	\end{tabular}
	\vspace{2mm}
	\caption{Comparison of our results against state-of-the-art batch and incremental methods}
	\label{table:compare_other}
\end{table}

\begin{figure*}[h]
	\centering
	\begin{tabular}{ccc}
		\includegraphics[scale=\ss]{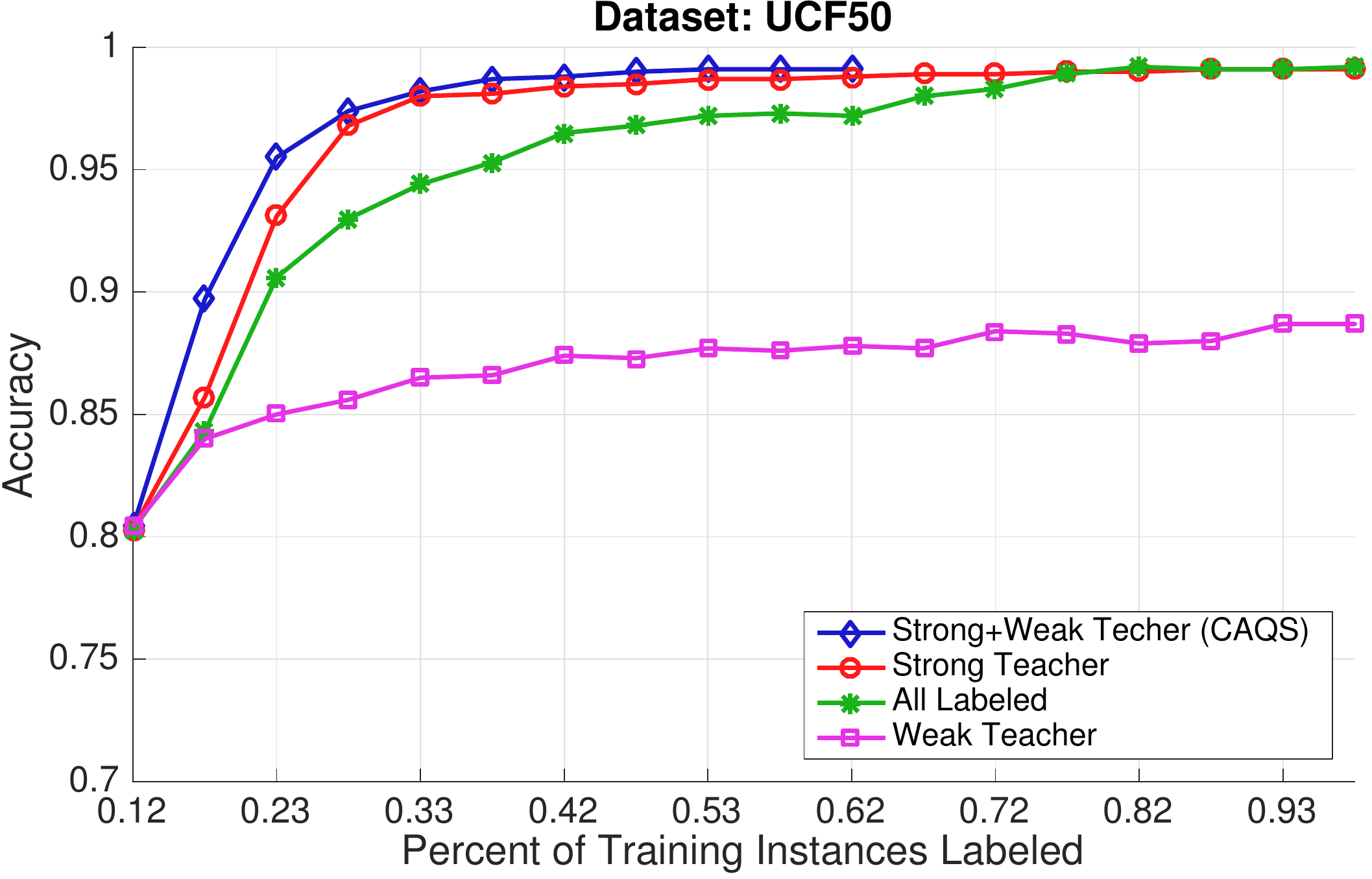} &
		\includegraphics[scale=\ss]{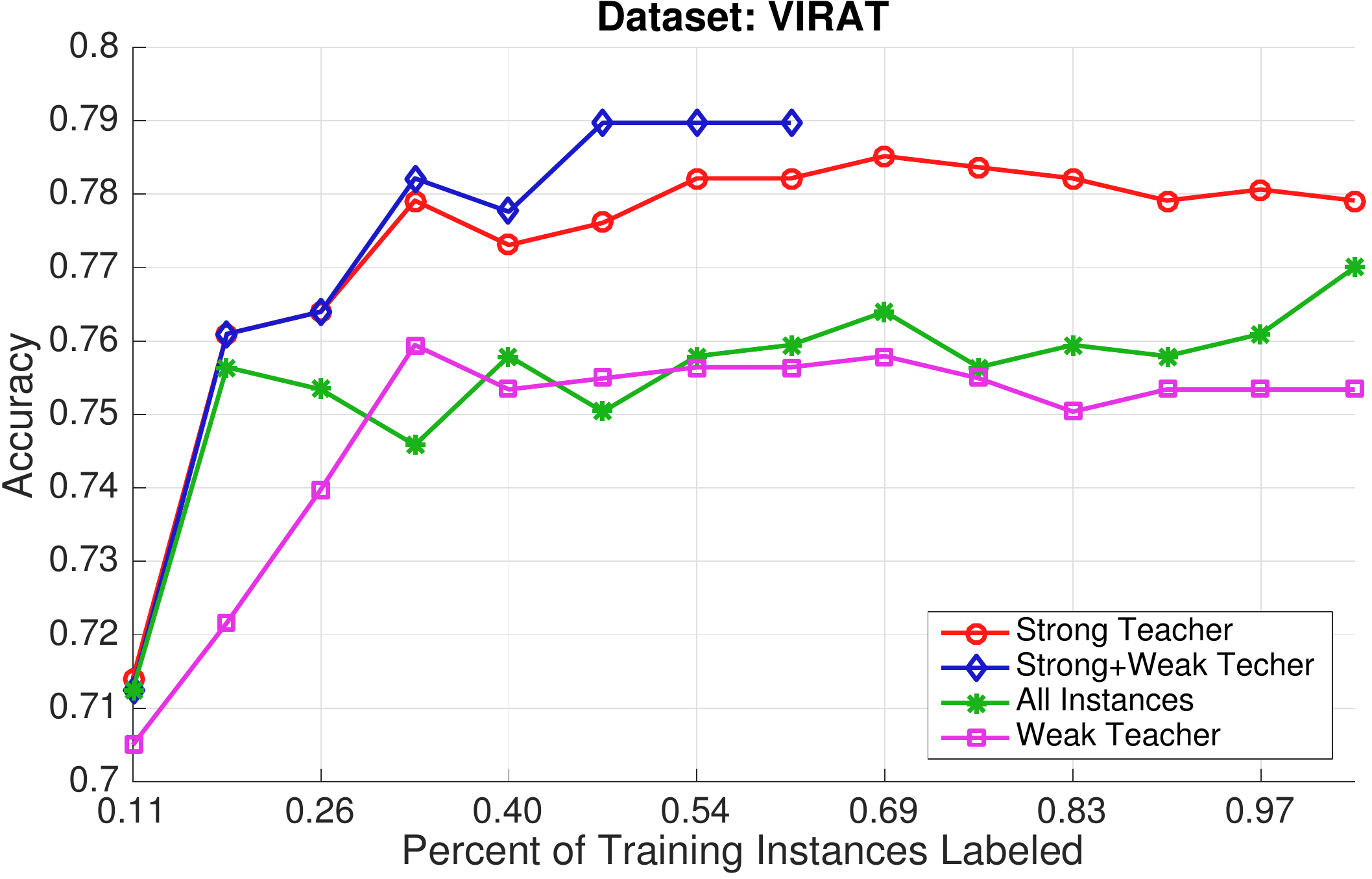} &
		\includegraphics[scale=\ss]{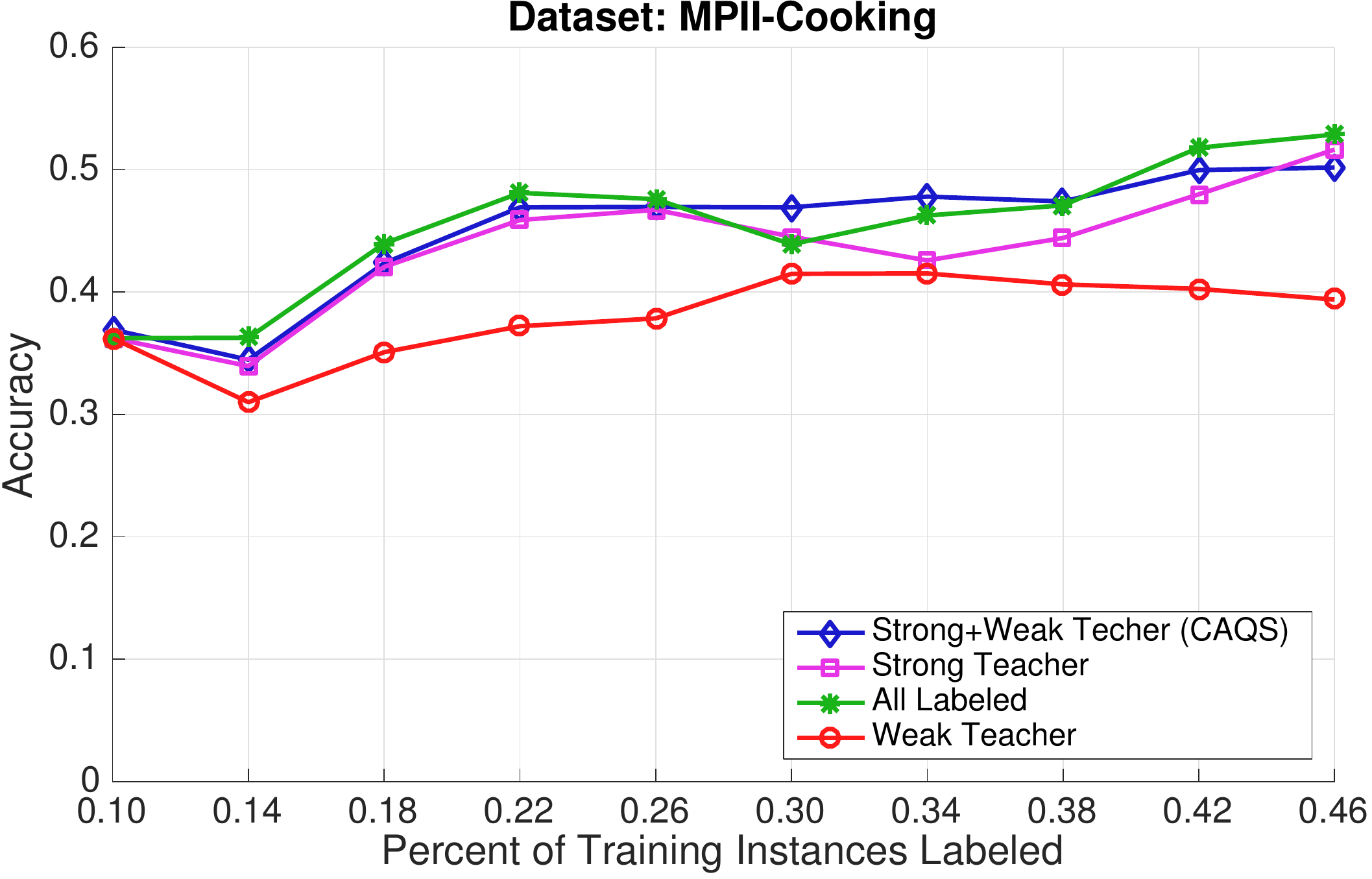} \\
		(a) & (b)  & (c) \\
	\end{tabular}
	\caption{Performance comparison among the four different variants of our proposed method. The X-axis represents the number of manually labeled training instances, whereas the Y-axis represents correct recognition accuracy on a set of unseen test instances. For a given value of X, all the method use same amount of manually labeled data, but the amount of labeled data can be different. Please see the text in Section \ref{expt:comp_self} for detailed explanation. Best view in color.}
	\label{plot:comps_self}
\end{figure*}

\subsection{Performance of Four Variants}
\label{expt:comp_self}
The plots in Figure~\ref{plot:comps_self} illustrate the comparison of our method's performance in four test cases, where we vary the use of the weak and strong teachers. These test cases are defined as follows. Weak teacher - for incremental training, we only use the highly confident labels provided by the model after the inference. No manually labeled instances are used in this test case. Strong teacher - we label a portion of the incoming instances manually. This portion is determined by the method described in Section \ref{sec:active}. Strong+Weak teacher - we use both of the above mentioned teachers. All instances - we manually label all the incoming instances to incrementally update the models. We can observe the following:
\begin{itemize}[leftmargin=*]
\item Performance of all of the test cases improves as more training instances are seen except for the weak teacher case. Weak Teacher only uses labels provided by the classifier, which are not always correct. These wrong labels of the training data lead to the classifier diverging over time. 
\item The strong+weak teacher uses around 40\% of manually labeled instances. However, its performance is very similar to the all-instance test case that uses 100\% manually labeled instances. This proves the efficiency of our method for selecting the most informative queries. In the plots, X-axis is the percentage of ``manually labeled" data. For a given value of X, all the method use same amount of ``manually labeled" data, but the amount of ``labeled" data can be different and it depends on the presence of weak teacher.

\item The performance of Strong+weak teacher and strong teacher are very similar. This indicates  that weakly labeled instances don't posses significant additional useful information for training because they are already confidently classified.

\item The performance with only weak teacher is not as good as the performance using strong teacher, because manual labels are provided only in the first batch. Afterwards, labels of the training instances are collected from the classifier, which are not correct always. As a result, its performance tends to diverge with time due to the training with noisy labels.  
\end{itemize}

\subsection{Experiment on AVA dataset}
\label{expt:ava}
{\bf Setup.} The experiment setup for AVA dataset is little bit different than the above setup due to the huge number of actions. In the above setup, we construct a graph with the entire training set and then iteratively perform active learning to select the most informative nodes in the graph. This process includes both message passing and inference in the graph. We use UGM \cite{schmidt2012ugm} for such task and unfortunately, it cannot handle such huge number of nodes and connections. In order to make the overall framework scalable, we consider one movie sequence out of 154 at a time and perform active learning with a pre-defined fraction. For example, at each step, we select $K = N_i * k$ number of nodes, where $k \in {[0,1]}$ and $N_i$ is the number of actions in movie $i$. The number of actions is heavily biased towards some action classes. For example, top ten activity classes contribute to about $85\%$ of the total activities, whereas more than fifty classes have less than 200 instances. This introduces both bias and noise in the model training and testing. We consider 28 activity types that have training and testing instances in the range of 200 and 10000.

{\bf Feature extraction.} As mentioned earlier, an action is 3 seconds long and only the middle frame is annotated with a bounding box. For simplicity, we assume that the action locations are spatially fixed and we spatially crop the actions from this 3 seconds of video using the bounding box. We extract 4096 dimensional C3D features from this cropped video and then, we use PCA to compress this 4096 dimensional features into 256 dimension for faster processing in the later steps.

\def \sv{0.26}
\begin{figure*}[h]
	\centering
	\begin{tabular}{ccc}
		\includegraphics[scale=\sv]{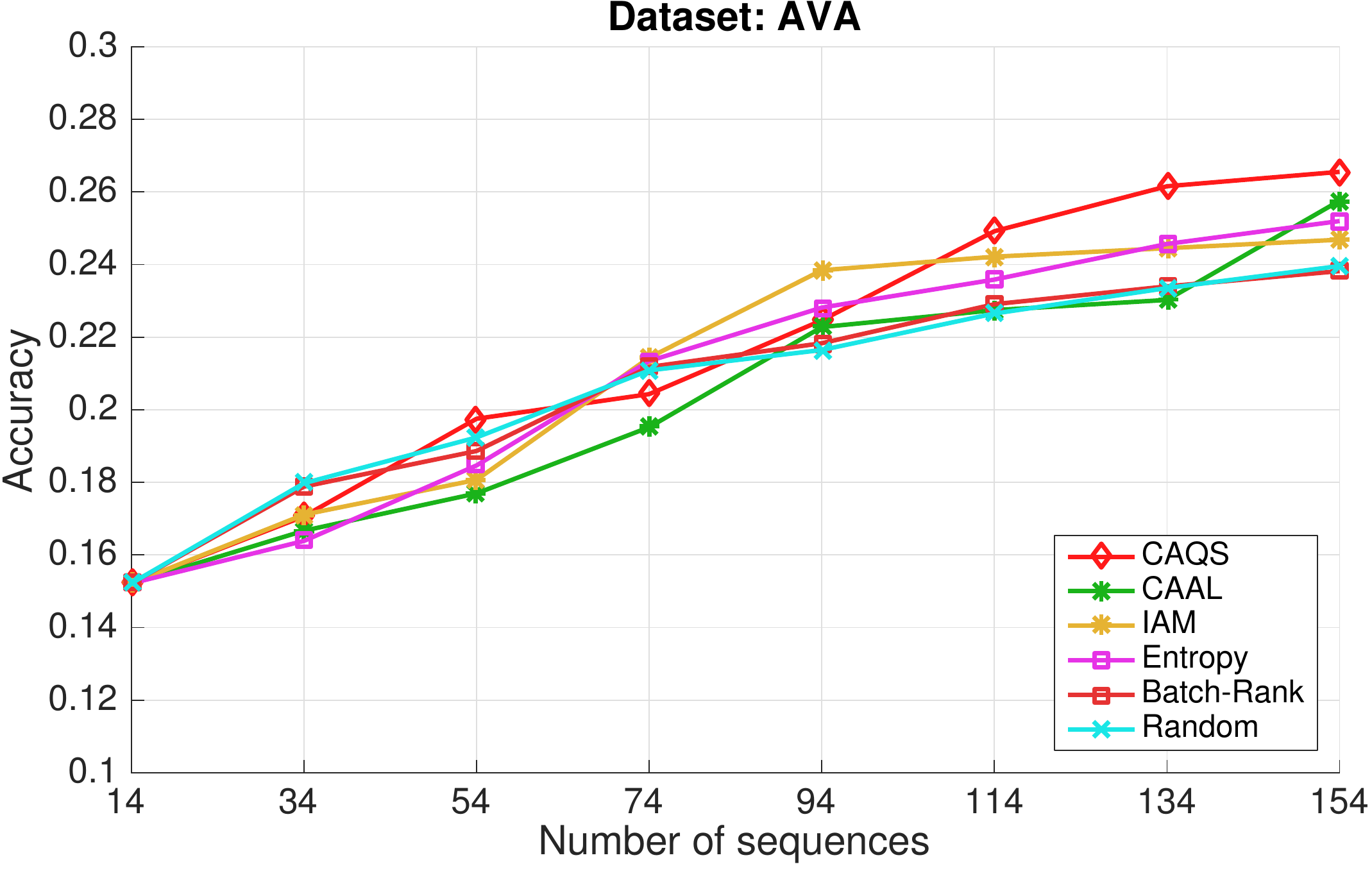}&
		\includegraphics[scale=\sv]{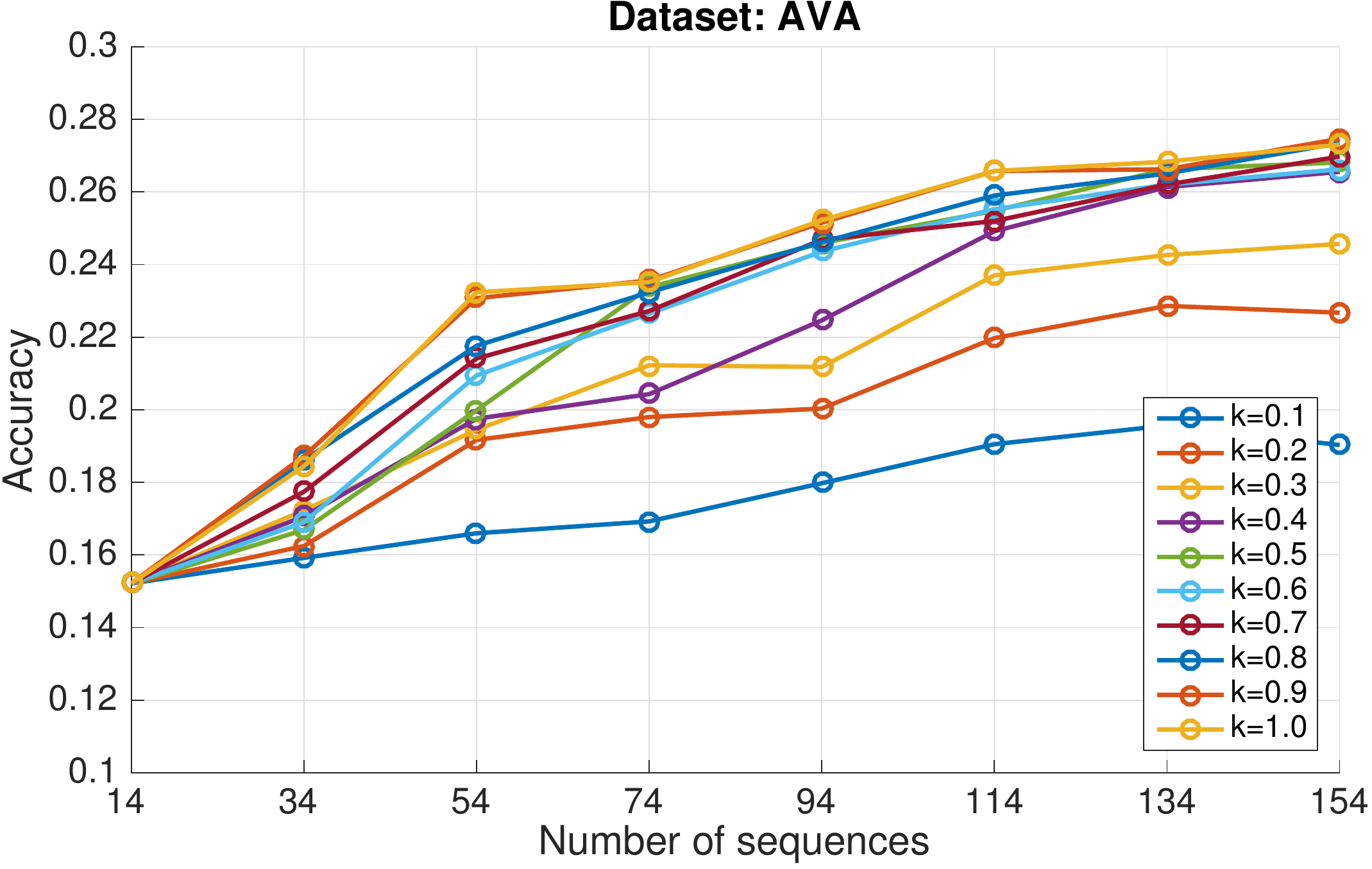} & 
		\includegraphics[scale=\sv]{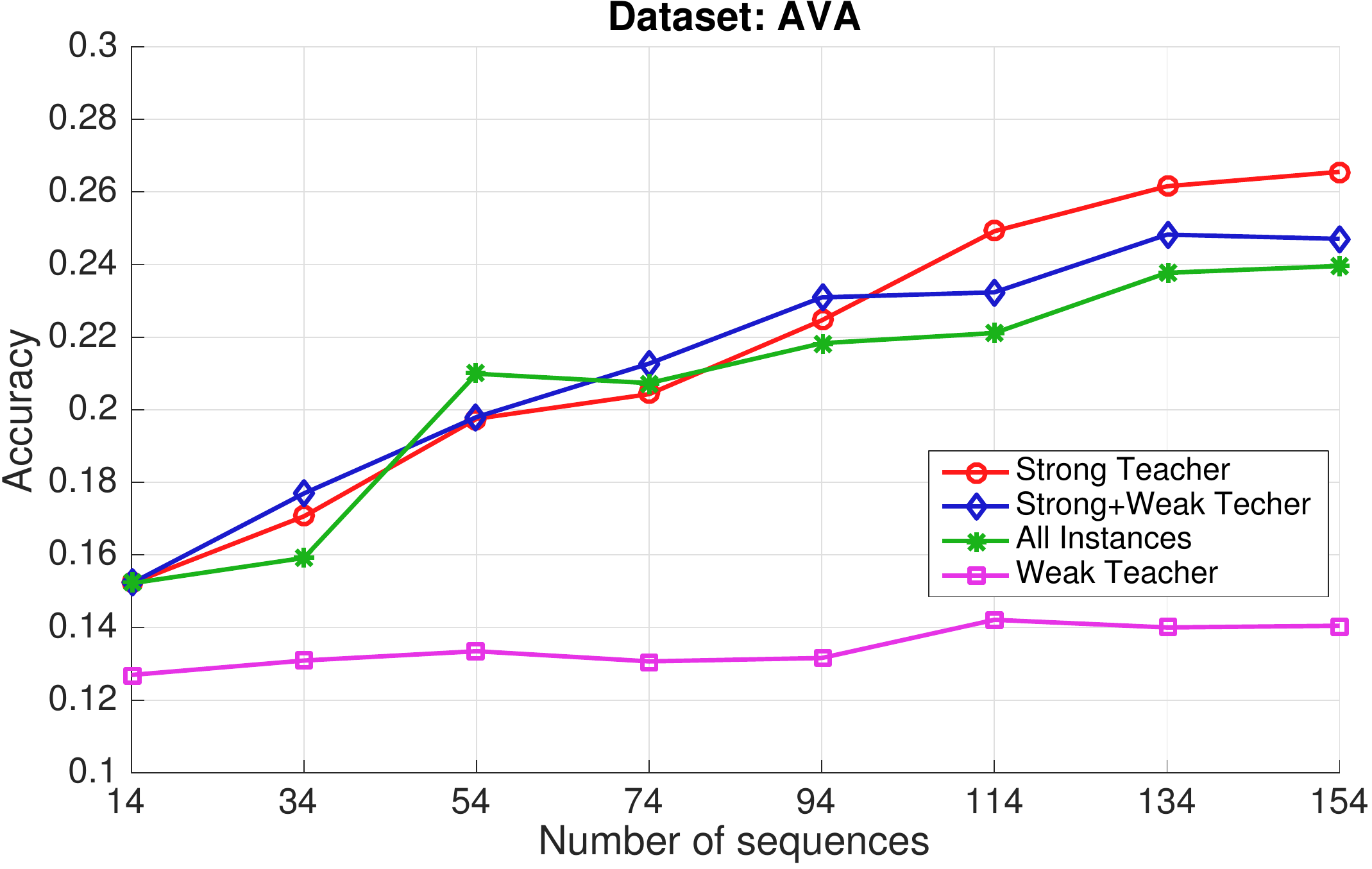}\\
		(a) & (b) & (c)
	\end{tabular}
	\caption{(a) Comparison with other state-of-the-art active learning methods. (b) Effect of number of query samples on the accuracy. (c) Performance comparison among the four variants. Please see the text in Section \ref{expt:ava} for detailed explanation. The X-axis is the sequence number of training videos. At each step, we use twenty training sequences to update the model and evaluate on the test set. In this experiment, for each sequence, we select forty percent of the most informative instances to be labeled manually.  The Y-axis is the accuracy of the prediction. Best viewed in color.}
	\label{plot:comps_ava}
\end{figure*}

{\bf Comparison with other active-learning methods.} 
In Fig. \ref{plot:comps_ava}(a), we compare our framework on AVA dataset with other state-of-the-art active learning methods as listed in Section \ref{expt:comp_al}. At the beginning, IAM performs well but our method, CAQS, outperforms every other method when all the video sequences are consumed. In this experiment, we use $k=0.4$. It means, we select the best forty percent of the instances to be labeled by the human. We use only the strong teacher for this dataset.

{\bf Effect of number of query samples.}
In Fig. \ref{plot:comps_ava}(b), we analyze the effect of the number of query samples on the prediction accuracy of our framework. Accuracy plots are pretty close to each other when $k$ is above $0.3$. It shows the robustness of the framework. Our framework can achieve the best performance using very few manually labeled instances.

{\bf Performance of four variants.}
In Fig. \ref{plot:comps_ava}(c), we show the performance of the four variants of the framework. The framework with only the strong teacher performs the best. The accuracy drops when we use weak teacher. This is because the overall accuracy is pretty low. As a result, the produced weak labels are also noisy sometimes, which makes the incrementally learned model noisy as well. The variant without any manual labels performs the worst as expected.

To the best of our knowledge, there is no research work yet to show action recognition accuracy on this dataset. The original paper \cite{AVA} shows mean average precision (MAP) on only activity detection results.

\subsection{Impact of Event Segmentation}
\label{sec:event-seg}
While the proposed method is agnostic to any event segmentation approach, in this section, we analyze the effect of an event segmentation algorithm (TCN) \cite{lea2017temporal}, vis-\`a-vis ground truth segmentation, on the performance of our proposed approach using 50Salads dataset \cite{stein2013combining}. We use TCN to obtain activity proposals along time without any label information attached to them. Any other method for generating activity proposals can also be used. We use the Encoder-Decoder (ED) variant among several TCN architectures. Our used ED-TCN has two layers in both encoder and decoder side and the size of the convolutional filter is set to 18. All other aspects of the network are kept same as in the original code. We use the provided features, which are extracted from spatialCNN \cite{lea2016segmental} network for every other fifth frame and has a dimension of 128. The outcome of the ED-TCN is frames labeled as activity or background as a proposal generation framework.

We use the segmentations/proposals provided by TCN as nodes of the graph in our model. Our goal is to select a subset of the nodes of this graph for manual labeling and model update thereafter. Note that initially, we do not have any class information about these segmentations. Once a subset of nodes/segmentations is selected for manual labeling, we obtain their labels as follows. We compare a proposal's temporal span with the actual ground truth in the dataset and assign the corresponding label if its temporal overlap with a certain activity is over $50\%$. Otherwise, we assign that proposal to belong to the background activity category.

\begin{figure}[h]
	\centering
	\includegraphics[scale=0.26]{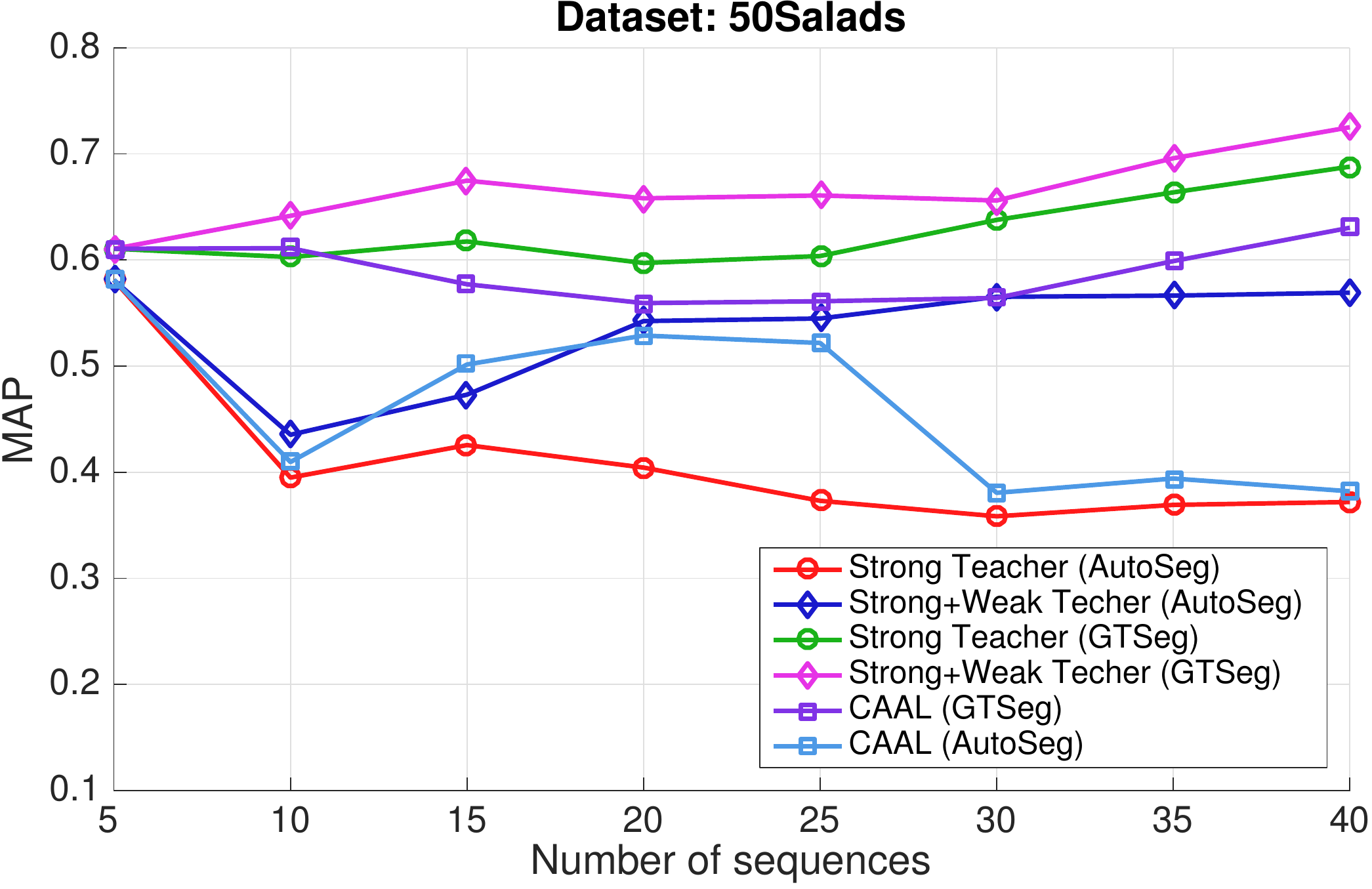}
	\caption{Comparison of the effect of different action segmentation methods. X-axis is the number of video sequences in the train set and Y-axis is the mean average precision. AutoSeg and GTSeg are automatic segmentation using TCN and truth segmentation respectively.}
	\label{plot:50salads}
\end{figure}

Fig. \ref{plot:50salads} shows the achieved results on both types of input segmentation with two variants of the proposed framework. 50Salads dataset \cite{stein2013combining} comes with five splits to define the train and test set. We use the first split in this experiment. There are forty training video sequences as shown in the X-axis and ten test video sequences. We report mean average precision (MAP) on the test set as the metric to compare as shown on the Y-axis. These results are achieved using forty percent labeling of the train set. 

Three plots in Fig. \ref{plot:50salads} correspond to ground truth segmentation and other three plots correspond to TCN segmentation. Four plots correspond to two variants of the proposed framework, i.e., only strong teacher and strong+weak teacher. Two of the plots correspond to the method CAAL. Ground truth segmentation achieved superior results as expected since segmentations from TCN are not as good as the ground truth. Strong teacher with the help of weak teacher achieves better results because labels obtained only from the strong teacher may be noisy due to imperfect segmentation. Proposed approach labeled as Strong+Weak Teacher performs the best among all compared methods for both ground truth segmentation, and the approach using \cite{lea2017temporal}, thus demonstrating that the proposed method outperforms others irrespective of the  activity segmentation. We achieved about $72.2\%$ (ground truth segmentation) and $57.8\%$ (TCN segmentation) MAP using only forty percent of the labeled data. Direct comparison with other approaches is not possible as we are not aware of existing methods on this dataset that use active learning.

\subsection{Comparison Against Deep Learning  Framework}
The goal of this section is to compare our proposed CRF based active learning method with a deep learning based active learning method. However, to the best of our knowledge, no such method exists to suit our experimental setup. So, we formulate an LSTM based active learning method similar to our proposed CRF based approach to capture context attributes. We use the first split of 50Salads dataset \cite{stein2013combining}, which contains 40 train and 10 test sequences. We divide the train set into batches and use the first batch to train our initial LSTM model, where we use five as the maximum sequence length with a stride of 1. The LSTM has one layer with 256 nodes and is trained using Adam optimizer with a cross entropy loss.

\begin{figure}[h]
	\centering
	\includegraphics[scale=0.26]{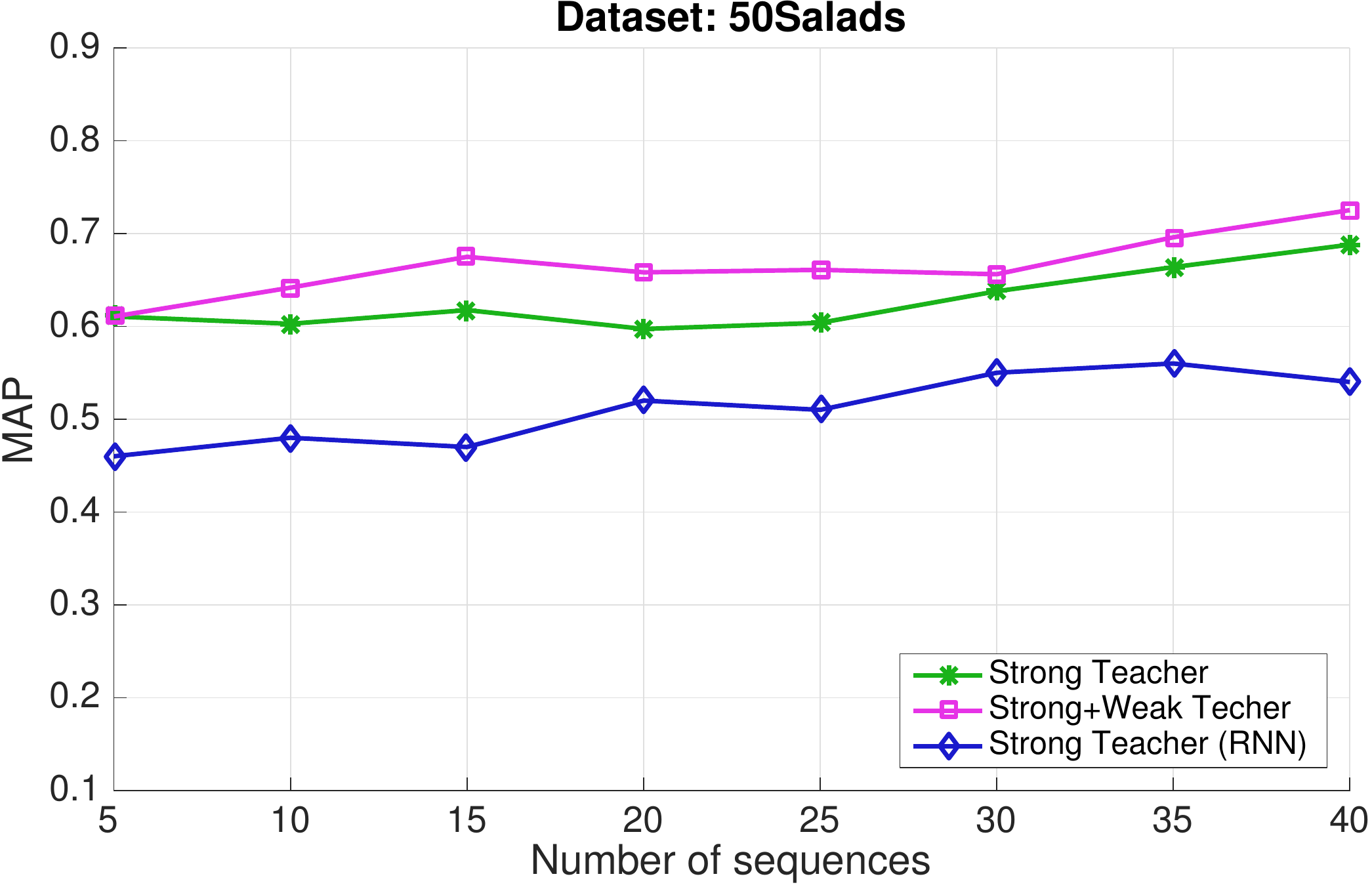}
	\caption{Comparison of the proposed context based active learning framework against the RNN based active learning framework using ground truth segmentation.}
	\label{plot:50salads_lstm}
\end{figure}

Using this initial LSTM model as the starting point, we perform active learning on the rest of the batches similar to our CRF based approach. We apply this LSTM on the next batch and use probability measure to apply weak and strong teacher. While this is a network with only one trainable layer, this could easily be extended to deeper network given we have sufficient data. For example, we use SpatialCNN network for feature extraction for 50Salads dataset. This network can be added with the LSTM network for end-to-end deep learning based active learning. However, only 1200 samples of 50Salads dataset was not sufficient for such task. In summary, we argue that for some tasks where the data is scarce and labeling is expensive, conventional CRF based method that can easily incorporate contextual domain knowledge much more efficiently than any deep learning based techniques. Also, our CRF based approach can be easily combined with transfer learning by using features from a network pre-trained with similar data. Fig. \ref{plot:50salads_lstm} compares the results of LSTM based approach against the proposed approach. X and Y axes are same as in Fig. \ref{plot:50salads}. As shown in the plots, LSTM based approach overfits and is not as good as the proposed framework to capture the context information among the actions.

\subsection{Parameter Sensitivity}
\label{sec:sens}		
\begin{figure*}[ht]
	\centering
	\begin{tabular}{ccc}
		\includegraphics[scale=0.26]{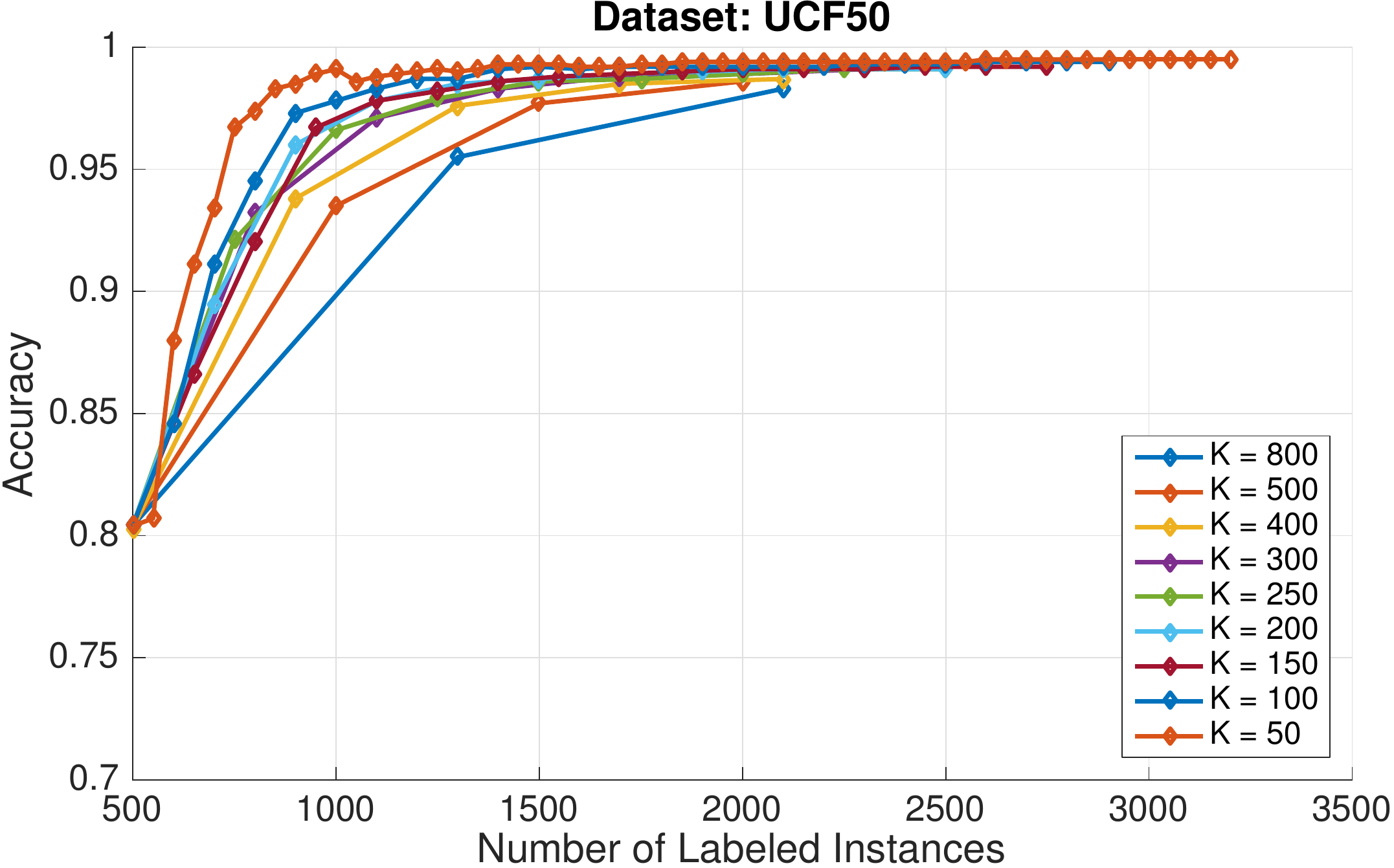} &
		\includegraphics[scale=0.26]{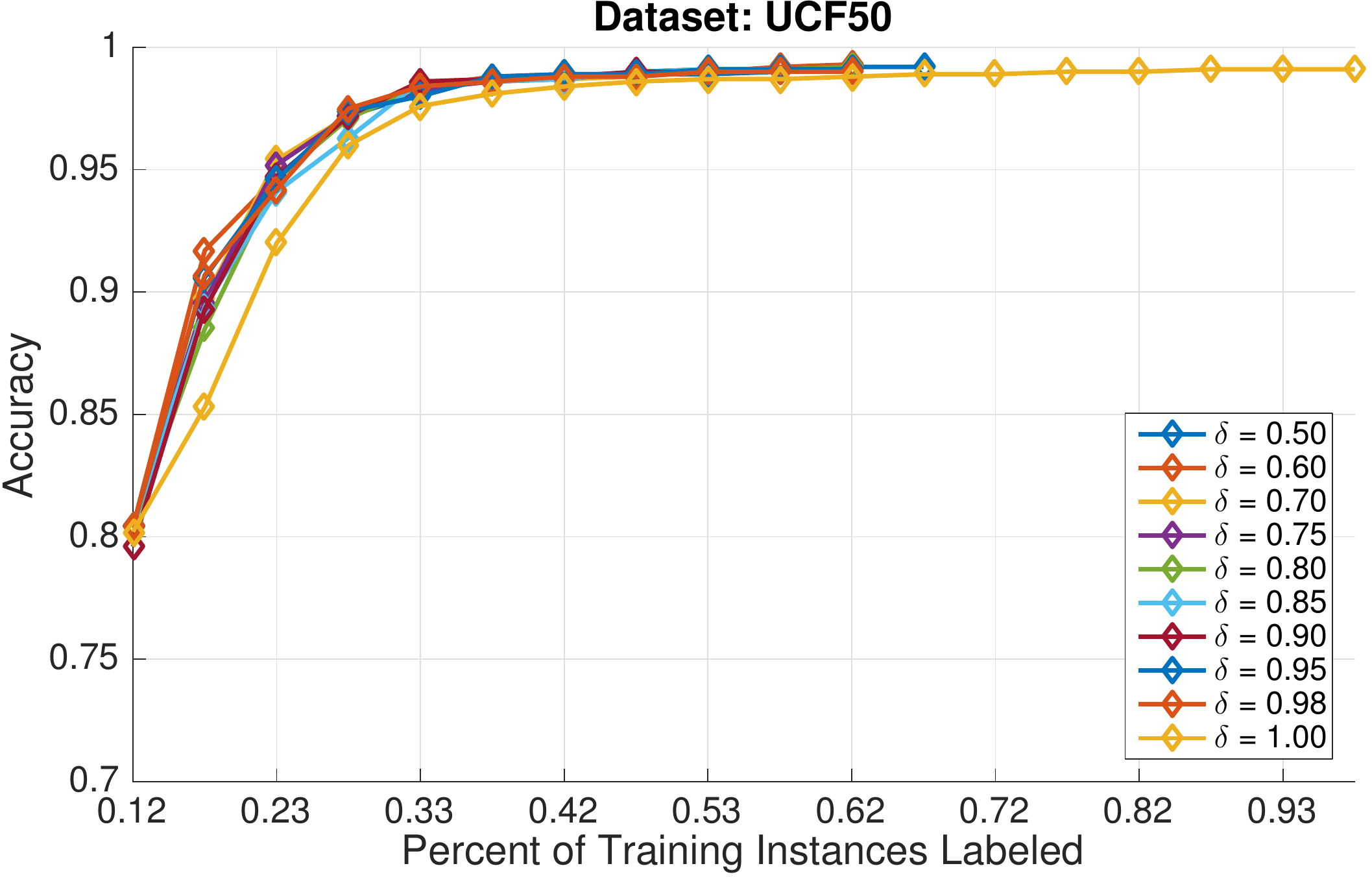} &
		\includegraphics[scale=0.26]{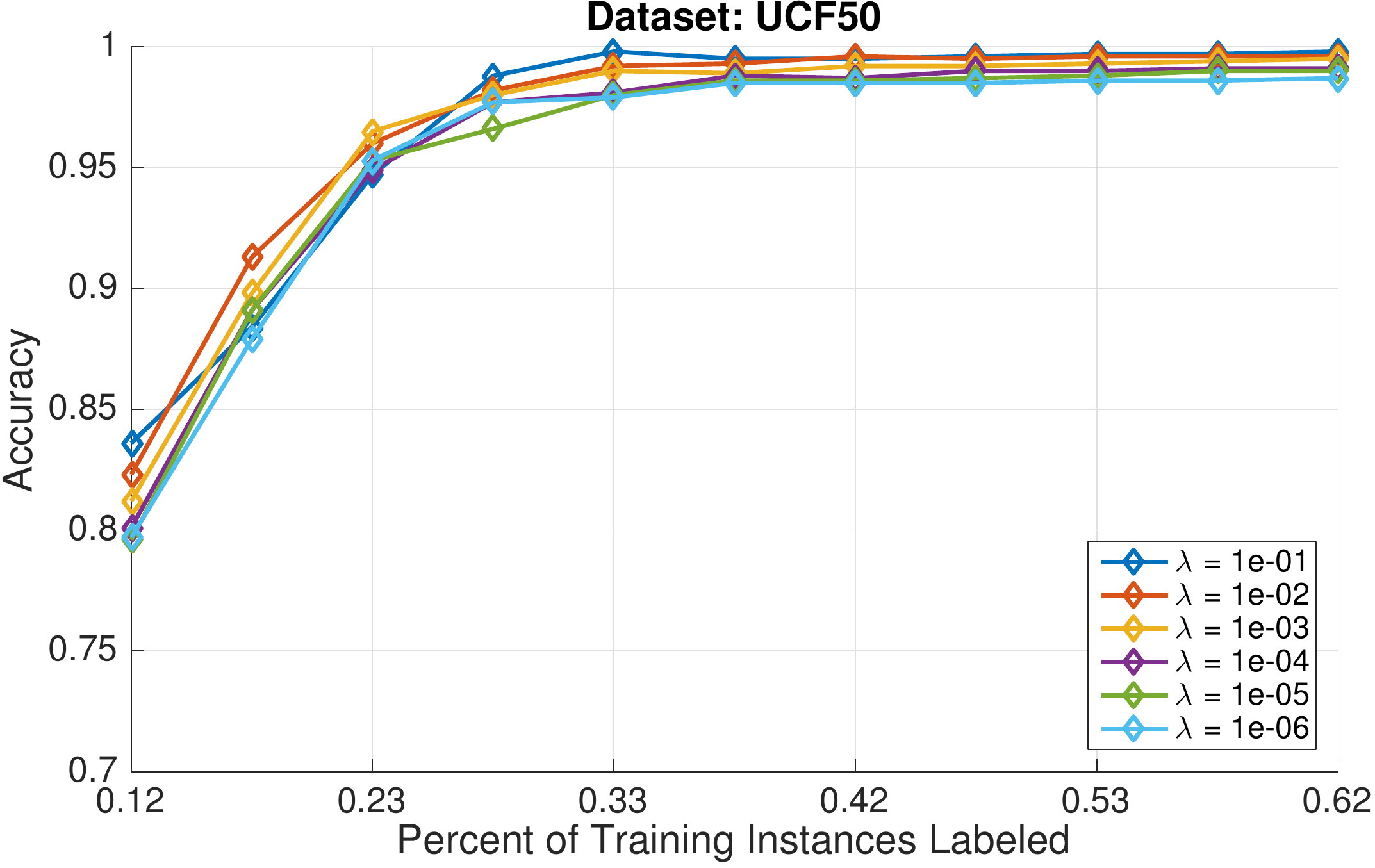}\\
		(a) & (b) & (c) \\
	\end{tabular}
	\caption{Plot (a), (b), and (c) show the sensitivity analysis of parameters $K, \delta, \text{ and } \lambda$ respectively on UCF50 dataset. The X-axis represents the number of manually labeled training instances, whereas the Y-axis represents correct recognition accuracy on a set of unseen test instances. Please see the text in Section \ref{sec:sens} for detailed explanation. Best view in color.}
	\label{plot:sensitivity}
\end{figure*}

Fig. \ref{plot:sensitivity}(a) shows the sensitivity analysis of the parameter $K$. At each iteration we select $K$ most informative instances from the training set that contains $m$ instances. We use them to train a classifier and apply this updated classifier on the test set. More specifically, At each iteration $i$, we select $K$ instances to be labeled by the strong teacher and few others instances (variable number say ($K^w_i$), depends on the threshold $\delta$) using weak teacher. At the beginning, we randomly select some instances (say $m_0$) to train the initial model. Next, at the first iteration, we have $m_0+K+K^w_1$ labeled instances to retrain the model. This is continued until we have any unlabeled training instances left. Lower values of $K$ provide better performance, as the selection of the queries becomes more fine grained. However, this makes the process more time consuming, because it increases the number of iterations needed, and training the classifier at each iteration is computationally expensive.

Fig. \ref{plot:sensitivity}(b) illustrates the sensitivity analysis of the parameter $\delta$. Our framework performs better for the higher values of $\delta$, where the framework uses very highly confident labels from the classifier to retrain it. For a lower value of $\delta$, it may be possible that some misclassified instances are used for retraining, which is the reason for inferior performance. The above two experiments use Strong+Weak Teacher active learning system. Fig. \ref{plot:sensitivity}(c) shows the sensitivity of the parameter $\lambda$. It has relatively lower impact on the UCF50 dataset.

\section{Conclusion}
\label{sec:conclusion}
We presented a continuous learning framework for context-aware activity recognition. We formulated an information-theoretic active learning technique that utilizes the contextual information among the activities and objects. We utilized entropy and mutual information of the nodes in active learning to account for the interrelationships between them. We also showed how to incrementally update the models using the newly labeled data. Finally, we presented experimental results to demonstrate the robustness of our method. 

Note that, our experimental setup does not include a real human. Whenever we need a label from the human we use the label from the ground truth. How to use the human efficiently given huge number of labels and classes is a different research problem on its own merit. One aspect of future work would be to understand the dynamics involved in human labeling, e.g., the amount of time to label, and how it interacts with the data ingestion and learning rate of the system. Another direction for future work would be to consider the localization problem (detection + recognition) in an active learning framework.

{
\bibliographystyle{IEEEtran}
\bibliography{CAQS_PAMI_2018}
}

~\small
\begin{wrapfigure}{r}{2.1cm}
	\centering
	\vspace{-3mm}
	\includegraphics[width=2.1cm]{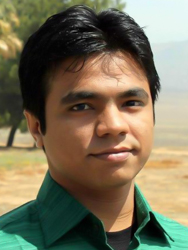}
\end{wrapfigure}
{\bf Mahmudul Hasan} graduated from UC Riverside with a Ph.D. in Computer Science in 2016. Previously he received his Bachelor's and Master's degree in Computer Science and Engineering from Bangladesh University of Engineering and Technology (BUET) in the year of 2009 and 2011 respectively. His broad research interest includes Computer Vision and Machine Learning with more focus on human action recognition, visual tracking, incremental learning, deep learning, anomaly detection, pose estimation, etc. He has served as a reviewer of many international journals and conferences.

~\small
\begin{wrapfigure}{r}{2.1cm}
	\centering
	\vspace{-3mm}
	\includegraphics[width=2.1cm]{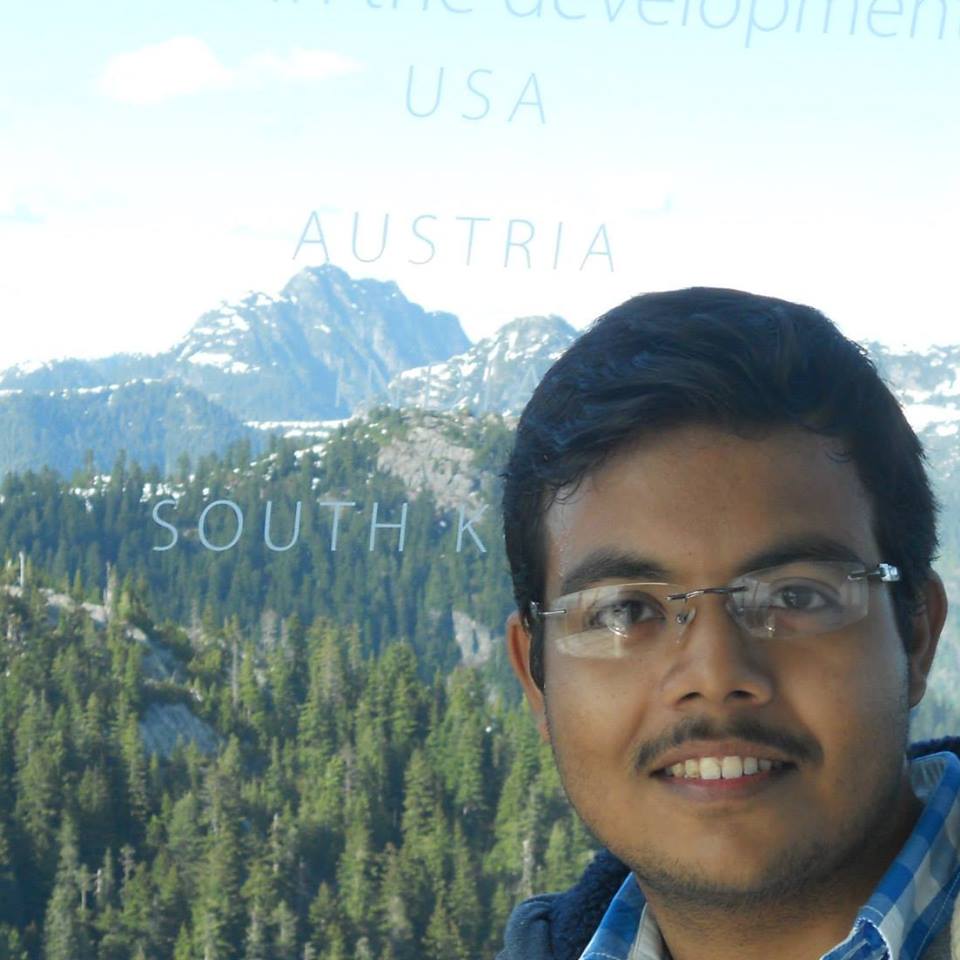}
\end{wrapfigure}
{\bf Sujoy Paul} received his Bachelor's degree in Electrical Engineering from Jadavpur University. Currently, he is pursuing his PhD degree in department of Electrical and Computer Engineering at University of California, Riverside. His broad research interest includes Computer Vision and Machine Learning with more focus on human action recognition, visual tracking, active learning, deep learning, etc.

~
\begin{wrapfigure}{r}{2.1cm}
	\centering
	\vspace{-3mm}
	\includegraphics[width=2.2cm]{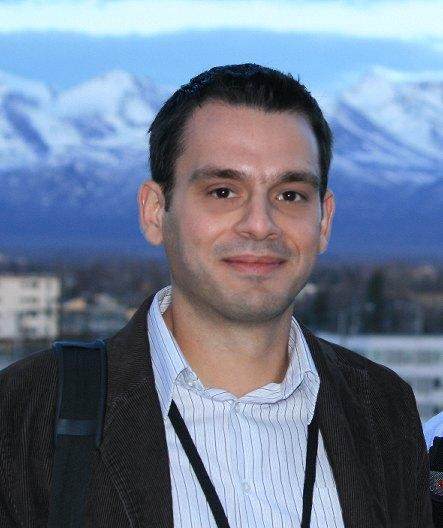}
\end{wrapfigure}
{\bf Anastasios I. Mourikis} received the Diploma in electrical engineering from the University of Patras, Patras, Greece, in 2003, and the Ph.D. degree in computer science from the University of Minnesota, Twin Cities, in 2008. He is currently an Assistant Professor in the Department of Electrical Engineering at the University of California, Riverside (UCR). His research interests lie in the areas of vision-aided inertial navigation, resource-adaptive estimation algorithms, distributed estimation in mobile sensor networks, simultaneous localization and mapping, and structure from motion. Dr. Mourikis has been awarded the 2013 National Science Foundation (NSF) CAREER Award, the 2011 Hellman Faculty Fellowship Award, and is a co-recipient of the IEEE Transactions on Robotics 2009 Best Paper Award (King-Sun Fu Memorial Award).

~
\begin{wrapfigure}{r}{2.1cm}
	\centering
	\vspace{-3mm}
	\includegraphics[width=2.2cm]{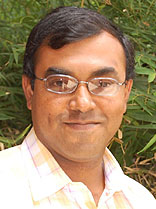}
\end{wrapfigure}
{\bf Amit K. Roy-Chowdhury} received the Bachelor's degree in electrical engineering from Jadavpur University, Calcutta, India, the Master's degree in systems science and automation from the Indian Institute of Science, Bangalore, India, and the Ph.D. degree in electrical engineering from the University of Maryland, College Park. He is a Professor of electrical engineering and a Cooperating Faculty in the Department of Computer Science, University of California, Riverside. His broad research interests include the areas of image processing and analysis, computer vision, and video communications and statistical methods for signal analysis. His current research projects include intelligent camera networks, wide-area scene analysis, motion analysis in video, activity recognition and search, video-based biometrics (face and gait), biological video analysis, and distributed video compression. He is a coauthor of two books Camera Networks: The Acquisition and Analysis of Videos over Wide Areas and Recognition of Humans and Their Activities Using Video. He is the editor of the book Distributed Video Sensor Networks. He has been on the organizing and program committees of multiple computer vision and image processing conferences and is serving on the editorial boards of multiple journals.

\end{document}


\title{Context-Aware Query Selection for Active Learning in Event Recognition}

\author{Mahmudul Hasan, Sujoy Paul, Anastasios I. Mourikis, and Amit K. Roy-Chowdhury\\
	University of California, Riverside \\
	{\tt\small \{mhasa004@, spaul003@, mourikis@ee., amitrc@ee.\}ucr.edu}}

\markboth{Journal of Pattern Analysis and Machine Intelligence,~Vol.~*, No.~*, Month~Year}{}%

\maketitle

\IEEEdisplaynontitleabstractindextext

\IEEEpeerreviewmaketitle

\section*{Appendix}

\begin{itemize}
	\item[A.] Proof of the statement (Section \ref{sec:obj-func-deriv}) that the manually labeled subgraph has zero entropy and mutual information.
	\item[B.] Proof of the statement (Section \ref{sec:opt-obj-func}) that the matrix $\boldsymbol Q$ is not positive semi-definite.
	\item[C.] Proof of the statement (Section \ref{sec:opt-obj-func}) that the addition of a suitable multiple of the identity to $\boldsymbol Q$ makes the objective function convex.
\end{itemize}

\subsection*{Appendix A}

Let us consider that node $a_i$ is manually labeled and then inference is performed on the graph conditioned on the obtained label of node $a_i$. Then the following is true:

\begin{itemize}
	\item As we consider that the label provided by human is correct, after labeling node $a_i$, the uncertainty associated with node $a_i$ drops to zero. Thus $\mathcal{H}(a_i)=0$.
	\item We know that $I(a_j;a_i)=H(a_i)+H(a_j)-H(a_i,a_j)$. Following the same argument as above, $H(a_i,a_j)=H(a_j)$, if $a_i$ is manually labeled. Thus $I(a_j;a_i)=H(a_i)+H(a_j)-H(a_j) = 0$.
\end{itemize}

To sum up, if a set of nodes is manually labeled and then we perform inference on the graph, the entropy of that set of nodes and the mutual information of all the edges linked to that set of nodes drops to zero. Thus,
\begin{align}
&\mathcal{H}(V^L_a) - \sum_{\substack{(i,j) \in E_a \\i \in V^L_a, j \in V^{NL}_a}} \mathcal{I}(a_j;a_i) \nonumber \\
&= \sum_{i \in V^L_a}\mathcal{H}(a_i)-\sum_{(i,j) \in E^L_a}\mathcal{I}(a_j;a_i) - \sum_{\substack{(i,j) \in E_a \\i \in V^L_a, j \in V^{NL}_a}} \mathcal{I}(a_j;a_i) = 0
\end{align}

\subsection*{Appendix B}
The matrix $\boldsymbol{Q}$ in Eqn . \ref{opt1} is not positive semi-definite (except when $\boldsymbol{Q}=\mathbf 0$). This is because $\boldsymbol{Q} \triangleq -\boldsymbol{M}$, where $\boldsymbol{M}$ is a matrix with diagonal elements equal to zero (recall that $\boldsymbol{M}(i,j)=0$ if $(a_i,a_j) \notin E_a$, and the graph does not contain self loops).

It is known that $Tr(\boldsymbol{M}) = 0 = \sum_{i=1}^N \lambda_i$ where $\lambda_i$s represent the eigenvalues of $\boldsymbol{M}$. The sum of all eigenvalues being equal to zero implies two possibilities:

\begin{itemize}
	\item $\boldsymbol M=\mathbf 0$. This is trivial, and if this is the case, then the optimization problem in Eqn. \ref{opt1} becomes much simpler. We neglect this condition.
	\item $\boldsymbol M \neq \mathbf 0$. In this case, both positive and negative eigenvalues will exist, and thus $\boldsymbol M$ (and $\boldsymbol Q$ will be indefinite.
\end{itemize}

Thus $\boldsymbol{Q}$ is not positive semi-definite except when no inter-activity edges exist in the graph, which is trivial.

\subsection*{Appendix C}
The value of $\gamma$ should be such that $\overline{\boldsymbol{Q}}=\boldsymbol{Q}+\gamma I$ is positive semi-definite. It is known that a diagonally dominant matrix having non-negative diagonal elements  is positive semi-definite. We therefore choose $\gamma$ so that the following condition holds:

\begin{align}
\gamma &\geq \sum_{j=1}^N |\boldsymbol{Q}(i,j)| =  \sum_{j=1}^N |\boldsymbol{M}(i,j)|   \quad \quad  \forall i \in {1,\dots,N} \nonumber \\
& \Rightarrow \gamma \geq \max_i{\sum_{j=1}^N |\boldsymbol{M}(i,j)|} = max\{|\boldsymbol M|\boldsymbol{1}\}
\end{align}
where the maximum is computed over all the elements of the vector $|\boldsymbol M|\boldsymbol{1}$, with $|\cdot|$ being the element-wise absolute-value operator.